\pdfoutput=1212112

\documentclass[11pt]{article}

\usepackage[]{acl}

\usepackage{times}
\usepackage{latexsym}

\usepackage[T1]{fontenc}

\usepackage[utf8]{inputenc}

\usepackage{microtype}

\usepackage{inconsolata}

\usepackage{times}
\usepackage{latexsym}
\usepackage{amsthm,amsmath,amssymb}
\usepackage{multirow}
\usepackage{array}
\usepackage{mathrsfs}
\usepackage{amsmath}
\usepackage{amsfonts}
\usepackage{graphicx}
\usepackage{booktabs}
\usepackage{algorithm}
\usepackage{algorithmicx}
\usepackage{algpseudocode}
\usepackage{CJKutf8}
\usepackage{xcolor}

\title{F-Eval: Assessing Fundamental Abilities with Refined Evaluation Methods}

\author{Yu Sun\textsuperscript{1,2}, Keyu Chen\textsuperscript{1,2}, Shujie Wang\textsuperscript{1,2}, Peiji Li\textsuperscript{1,2}, Qipeng Guo\textsuperscript{1}, Hang Yan\textsuperscript{1,3}\thanks{Corresponding Author}, \\
\textbf{Xipeng Qiu\textsuperscript{2}, Xuanjing Huang\textsuperscript{2}, Dahua Lin\textsuperscript{1,3}}\\
  \textsuperscript{1}Shanghai AI Laboratory, \textsuperscript{2}School of Computer Science, Fudan University,\\ \textsuperscript{3}The Chinese University of Hong Kong\\
  \texttt{\{yusun21,kychen22\}@m.fudan.edu.cn},\\ \texttt{\{wangshujie,guoqipeng,yanhang,lindahua\}@pjlab.org.cn}\\
  \texttt{\{xpqiu,xjhuang\}@fudan.edu.cn}\\}

\begin{document}
\maketitle
\begin{abstract}

Large language models (LLMs) garner significant attention for their unprecedented performance, leading to an increasing number of researches evaluating LLMs. However, these evaluation benchmarks are limited to assessing the instruction-following capabilities, overlooking the fundamental abilities that emerge during the pre-training stage. Previous subjective evaluation methods mainly rely on scoring by API models. However, in the absence of references, large models have shown limited ability to discern subtle differences. To bridge the gap, we propose F-Eval, a bilingual evaluation benchmark to evaluate the fundamental abilities, including expression, commonsense and logic. The tasks in F-Eval include multi-choice objective tasks, open-ended objective tasks, reference-based subjective tasks and reference-free subjective tasks. For reference-free subjective tasks, we devise new evaluation methods, serving as alternatives to scoring by API models. We conduct evaluations on 13 advanced LLMs. Results show that our evaluation methods show higher correlation coefficients and larger distinction than other evaluators. Additionally, we discuss the influence of different model sizes, dimensions, and normalization methods. We anticipate that F-Eval will facilitate the study of LLMs' fundamental abilities.~\footnote{The code and dataset is available in \url{https://github.com/OpenLMLab/F-Eval}.}

\end{abstract}

\section{Introduction}
Large language models (LLMs)~\cite{zeng2022glm,chatgpt,touvron2023llama,baichuan2023baichuan2,qwen,2023internlm} are gaining increasing popularity in the field of Natural Language Processing (NLP), owing to their commendable performance in various applications. As the development of LLMs is blooming, there has been a surge of evaluation work to probe the diverse capabilities of LLMs.

\begin{figure}[t]
  \centering
  \includegraphics[width=\columnwidth]{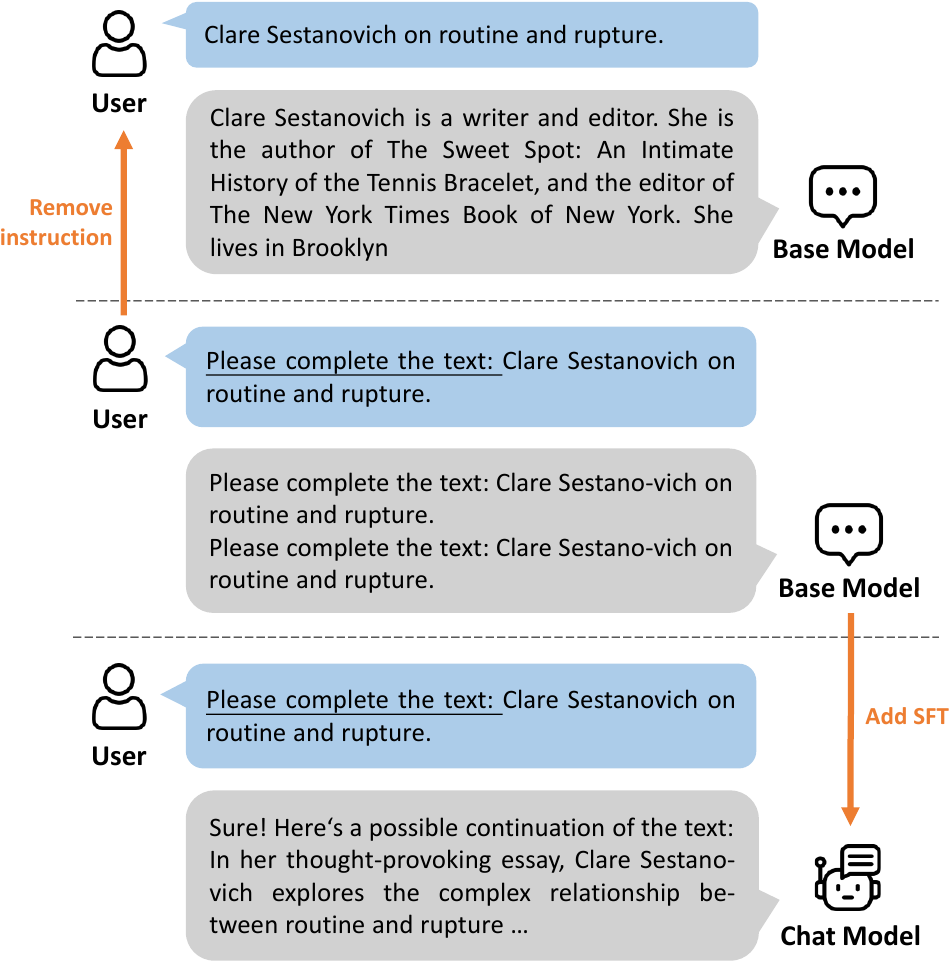}
  \caption{Prompts with instructions limit the capabilities of the base model (middle), which can be resolved either by removing instructions (upper) or by further SFT on the base model (lower).} \label{fig:intro}
\end{figure}

\begin{figure*}[ht]
    \centering
  \includegraphics[width=0.9\textwidth]{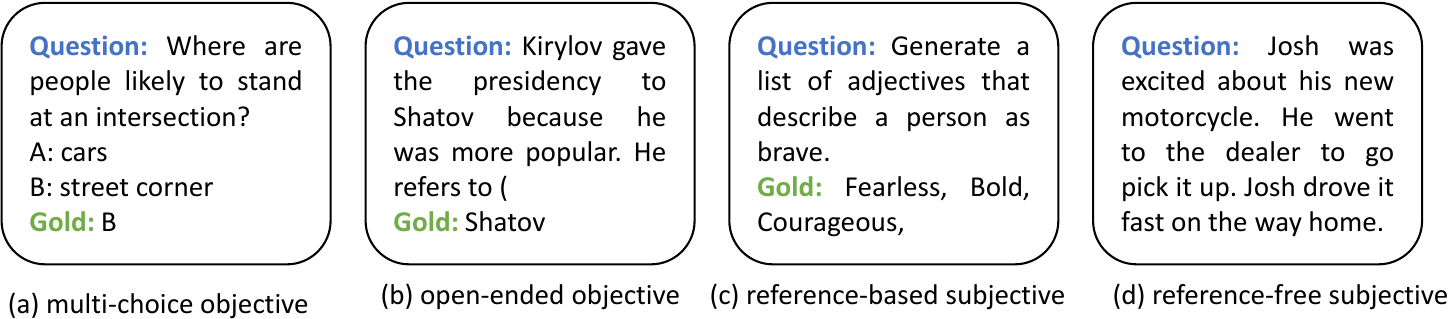}
  \caption{The examples of each data format.} \label{fig:data_format}
\end{figure*}

Objective benchmarks~\cite{hendryckstest2021,DBLP:journals/corr/abs-2206-04615,huang2023ceval,DBLP:journals/corr/abs-2306-09212} primarily focus on the model's problem-solving abilities across different subjects, without considering alignment with human in real-world scenarios. Consequently, a series of subjective evaluation efforts~\cite{alpaca_eval,naturalinstructions,zheng2023judging,liu2023alignbench} emerge, shifting the focus to instruction-following and conversational capabilities of LLMs. However, these benchmarks are based on the assumption that LLMs can understand complex instructions and questions, which only emerges after the Supervised Fine-Tuning (SFT) stage. The example in Figure \ref{fig:intro} demonstrates that the ability of base models before SFT is susceptible to instructions. Currently, benchmarks mainly focus on the evaluation of chat models after SFT (lower section), there is still a lack of benchmarks that assess the basic abilities possessed in LLMs during their pre-training stage~(upper section). In our paper, we define this basic ability, which does not rely on additional fine-tuning and is required during the pre-training stage of LLMs, as "fundamental ability".

In addition, subjective evaluations require the generations of LLMs to be consistent with human experience. Current subjective evaluations heavily rely on scoring by API models, such as GPT4.0~\cite{openai2023gpt4}. \citet{zheng2023judging} observe that LLMs are good evaluators when scoring with references. However, they find that without references, LLMs have limited capability in discerning the quality of outcomes. This may be due to the fact that LLMs make judgments based on their internal knowledge, which is chosen randomly, leading to unstable scoring results and low distinction.

To bridge the gaps in evaluation focuses and methods, we propose F-Eval, the first evaluation benchmark to thoroughly assess LLMs' fundamental abilities, which is applicable to both base models and chat models. The datasets in our benchmark consist of 2211 instances in both English and Chinese with 3 dimensions, including expression, commonsense, and logic. We design a total of 15 sub-datasets, encompassing formats such as multi-choice objective tasks, open-ended objective tasks, reference-based subjective tasks, and reference-free subjective tasks. We show an example for each data format in Figure~\ref{fig:data_format}. The composition of the dataset is shown in Figure \ref{fig:dataset}. For objective questions, we use accuracy as metrics. For reference-based subjective tasks, we prompt GPT4.0\footnote{We use gpt4-preview-1106 version for GPT4.0.} as the evaluator. As for reference-free subjective tasks, we design more stable and distinctive evaluation methods to replace scoring by API models. The evaluation methods corresponding to each sub-dataset are listed in Table \ref{tab:evaluation_methods}.

We conduct experiments to evaluate 13 advanced LLMs on F-Eval. The results reveal that open-source models still maintain a large gap to GPT4.0, highlighting a considerable room for improvement of LLMs. Our experiments show that F-Eval outperforms other baselines in terms of correlation with human judgements. Meanwhile, the evaluation methods designed for reference-free subjective tasks have larger distinction than LLM scoring. To delve into the performance, detailed discussions uncover the impact of model size on capabilities and the imbalance ability of different models across three dimensions. Additionally, we have also demonstrated the superiority of our specially designed method for normalizing results.

To summarize, our contributions are as follows:

\begin{itemize}
    \item We introduce F-Eval, the first comprehensive benchmark to evaluate the fundamental ability of LLMs. The data in the benchmark is divided into 15 sub-datasets across 3 dimensions. 
    \item To employ suitable evaluation methods for each sub-dataset, we use 4 categories of evaluation methods. Among these, we specifically devise new methods for reference-free subjective tasks, serving as an alternative to scoring by API models. Our experiments have shown that our evaluation methods perform well in terms of consistency with human evaluations and in distinguishing the outputs.
    \item We comprehensively discuss the performance of LLMs within different model sizes, dimensions and normalization methods, expecting to shed a light on the improvement on fundamental ability for further LLM researches.
\end{itemize}

\section{Related Work}

Recent advancements in large language models have attracted significant interest, with the depth and breadth of evaluation work consistently expanding. Broadly speaking, these evaluations can be classified into two distinct categories: objective evaluations and subjective evaluations.

Objective evaluations typically adopt formats such as multiple-choice queries and some open-ended questions with definitive responses. A large proportion of the multiple-choice benchmarks~\cite{hendryckstest2021,huang2023ceval,DBLP:journals/corr/abs-2306-09212} are task-oriented, primarily assessing the model's question-answering capabilities. Open-ended questions frequently encompass a range of knowledge queries, as exemplified by the NaturalQuestions~\cite{47761} and TriviaQA~\cite{2017arXivtriviaqa}. Apart from knowledge, reasoning capabilities are often a key focus of evaluation, such as GSM8K~\cite{cobbe2021gsm8k} and TheoremQA~\cite{chen2023theoremqa}. However, a notable limitation of objective evaluations is their misalignment with human, leading to high scores but do not correlate with users' experience.

Subjective evaluations, on the other hand, aim to harmonize with human experiences, primarily gauging the ability to adhere to instructions and engage in dialogues. These evaluations typically employ a scoring system for LLMs APIs, such as the GPT4.0 and GPT3.5. There's a wealth of research in this domain, including AlignBench~\cite{liu2023alignbench}, which offers a comprehensive, multi-dimensional evaluation benchmark for Chinese LLM alignment, utilizing a rule-based language model for evaluation. AlpacaEval~\cite{dubois2023alpacafarm} provides a fully automated evaluation benchmark based on the LLM and employing GPT4.0 or Claude as automatic evaluators. The benchmark compares the target model's responses with those of GPT3.5 and calculates the win rate. Several studies~\cite{chia2023instructeval,liu-etal-2023-g,fu2023gptscore,chen2023phoenix} have also focused on how to utilize LLMs for scoring. However, current subjective evaluations heavily rely on references. In scenarios where references are not available, the quality of LLMs' results is limited, failing to accurately reflect the ability partial order of LLMs.
We propose a new evaluation dataset and evaluation method that can address these issues. It can examine both base models and chat models. At the same time, it allows subjective evaluations to have higher credibility and greater distinction when there is no reference.

\section{Benchmark}

To assess the fundamental capabilities of LLMs, we design F-Eval to examine the model's fundamental abilities from 3 dimensions and establish corresponding appropriate evaluation methods for each sub-dataset. 

\subsection{Data Collection}
\label{sec:dataset}

\begin{figure}[h]
  \centering
  \includegraphics[width=\columnwidth]{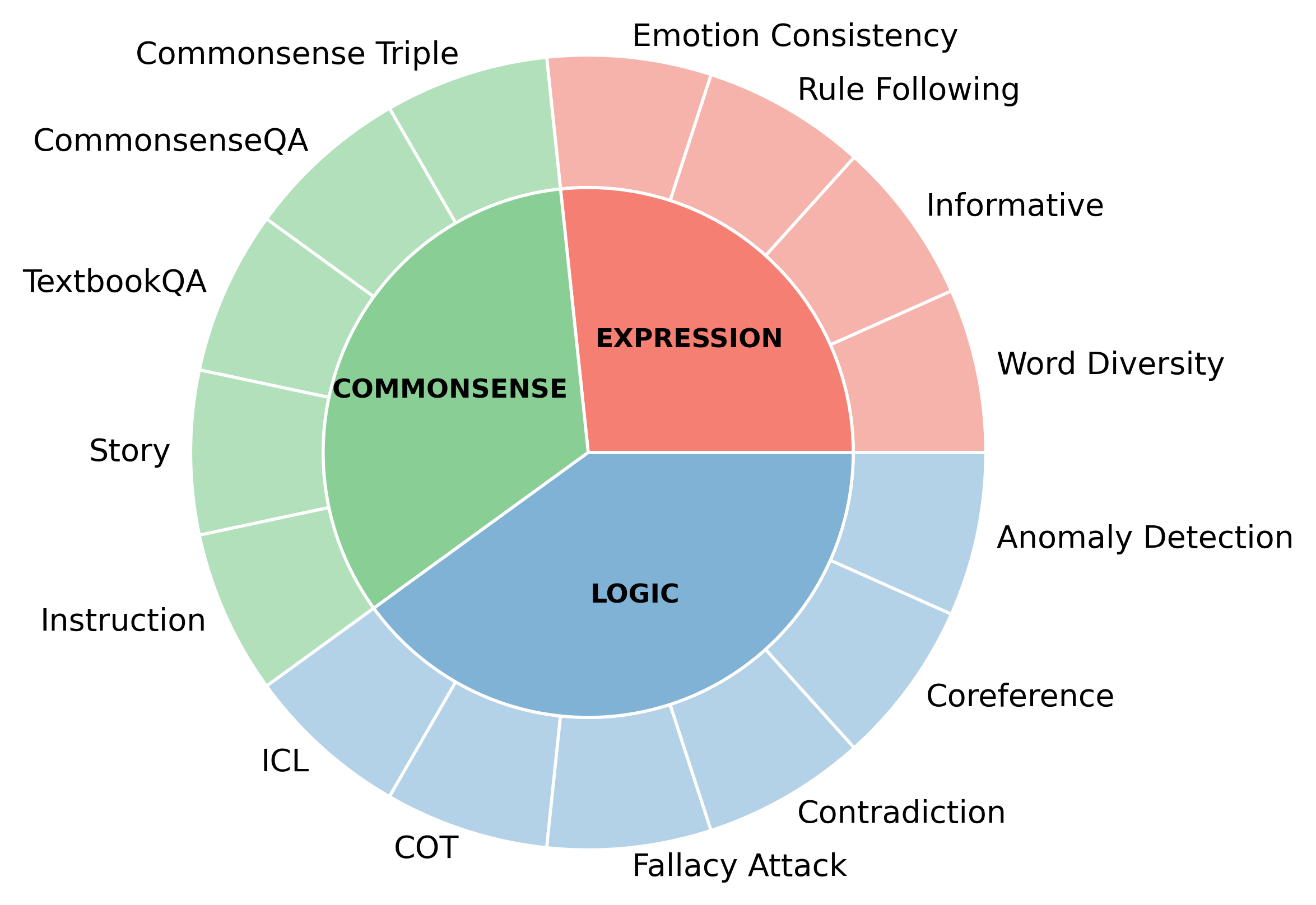}
  \caption{Overall composition of F-Eval.} \label{fig:dataset}
\end{figure}

Our dataset contains 15 sub-datasets with 2211 instances in both English and Chinese. The overall composition of our dataset is shown in Figure \ref{fig:dataset}. Each sub-dataset contains both English and Chinese data. Detailed descriptions and examples of each sub-dataset are shown in Appendix \ref{sec:appendix}.

\paragraph{Expression} To examine the quality of LLMs' generated texts, the first aspect to consider is the model's expressive ability. Sub-datasets in this dimension mainly investigates the diversity of words (Word Diversity), consistency in the quantity of information (Informative), consistency in writing format (Rule Following), and consistency in emotional style (Emotion Consistency) of the generated texts. 
Among them, Rule Following dataset is an open-ended objective tasks, while all other sub-datasets are reference-free subjective tasks.

\paragraph{Commonsense} In this part, our primary focus is on assessing the LLM's grasp of commonsense. On the one hand, to examine the awareness of LLMs on commonsense knowledge, we create three sub-datasets, Commonsense Triple, CommonsenseQA and TextbookQA, which directly ask questions about commonsense in various types. On the other hand, to verify whether the LLM can apply commonsense to make simple event predictions, we introduce two sub-datasets, Story and Instruction, to allow the LLM to select appropriate story endings and answer instructions based on commonsense, respectively. CommonsenseQA and Story are formed as multi-choice objective tasks, while others are reference-based subjective tasks.

\paragraph{Logic} As a language model, the logical abilities of LLMs can mainly be demonstrated in three aspects: induction, reasoning, and logical coherence. To evaluate whether LLMs can induce the requirements and forms of output from in-context examples, we construct a in-context learning (ICL) dataset. The ability naturally emerges as LLMs reach a certain scale. In the aspect of reasoning, LLMs are expected not only to infer correct answers based on valid reasoning chains (COT), but also to possess the ability to discern and correct fallacious reasoning chains using commonsense (Fallacy Attack). Finally, the generated texts should maintain logical consistency, such as avoiding contradictory statements (Contradiction), accurately identifying coreferences (Coreference), and recognizing incorrect coreferences (Anomaly Detection).

With only Fallacy Attack being built only by humans, 14 of 15 sub-datasets are automatically collected. Among them, Word Diversity, Informative, Rule Following and Contradiction are derived from the collection of online data, while others are derived from adaptations of existing datasets. More details about data sources are shown in Appendix \ref{sec:appendix}. To prevent LLMs from memorizing the existing examples, we adhere to the following two principles during the automatic collection of data. Firstly, for the data adapted from existing datasets, we change the data format and expression manually or by LLMs. Secondly, the online data we collect is mostly from documents post after June 2023. Additionally, to ensure the quality of our dataset, we thoroughly review and refine the instances which are uniformly answered incorrectly by all LLMs. Moreover, we analyze the accuracy distribution across multiple LLMs and adjust the number of examples with either too low or too high accuracy rates. This adjustment aims to make the score distribution as even as possible or to follow a normal distribution.

\subsection{Evaluation Methods}

The evaluation methods of the sub-datasets in F-Eval are listed in Table \ref{tab:evaluation_methods}. We give a brief introduction of each method as follows, more details are described in Appendix \ref{sec:appendix}.

\begin{table}[htb]
\renewcommand\arraystretch{1.1}
  \centering
  \small
  \setlength{\tabcolsep}{3pt}
  \tabcolsep=0.3cm
  \begin{tabular}{@{}lllll@{}}
    \toprule
        Evaluation Methods & Sub-Dataset      \\
        \midrule
        \multirow{4}{*}{Rule-based Evaluation} & Rule Following\\
        & ICL \\
        & COT \\
        & Coreference\\
        \midrule
        \multirow{3}{*}{Probability Evaluation} &  CommonsenseQA \\
         & Story \\
         & Anomaly Detection\\
         \midrule
        \multirow{4}{*}{Assistant-Tool Evaluation}  & Word Diversity \\  
        & Informative \\
        & Emotion Consistency \\
        & Contradiction\\
        \midrule
        \multirow{4}{*}{API Evaluation}  & Commonsense Triple \\
        & TextbookQA\\
        & Instruction\\
        & Fallacy Attack\\
         
          \bottomrule
  \end{tabular}
  \caption{Evaluation Methods.}
  \label{tab:evaluation_methods}
\end{table}

\paragraph{Rule-based Evaluation}

Rule-based evaluation method simply relies on the generation and the designed rules, which is applied on open-ended objective sub-datasets. ICL and Coreference require the prediction to exactly match the gold answer. As for Rule Following and COT, We design matching rules to determine whether the generation meets our requirements. We use accuracy as the metrics.

\paragraph{Probability Evaluation}

For multi-choice objective tasks, including CommonsenseQA, Story and Anomaly Detection, we follow previous work~\cite{DBLP:journals/corr/abs-2306-09212,huang2023ceval} to use probability of the entire text for evaluation. We select the option corresponding to the prompt with the highest probability as the prediction. As for non open-source models like GPT4.0 and GPT3.5\footnote{We use gpt-3.5-turbo-1106 version for GPT3.5.}, we prompt them to directly generate the option. Then we use accuracy as the metrics. 

\paragraph{Assistant-Tool Evaluation}

For reference-free subjective tasks, we leverage assistant tools as the evaluator instead of API models. On the one hand, we utilize dictionaries to evaluate the rarity and diversity of vocabulary in LLMs' generation for Word Diversity sub-dataset. On the other hand, we use assistant models to evaluate. The probability of LLMs can be regarded as the amount of information contained in a text~\cite{radford2018improving}. Therefore, when evaluating Informative sub-dataset, we use a judge model to calculate the difference of the probability between input and output, indicating the consistency of information. Additionally, the evaluation focus of some datasets within F-Eval aligns with the task orientation of traditional NLP models, such as Emotion Consistency and Contradiction sub-datasets. Therefore, we use them to identify the sentiment and contradiction. We regard the designed ratio as scores.

\paragraph{API Evaluation}

For reference-based subjective tasks, we follow \citet{zheng2023judging} to choose the best-performed GPT4.0 as the evaluator. Specifically, we follow AlignBench~\cite{liu2023alignbench} to design our evaluation prompt.

\subsection{Results Normalization}
\label{sec:norm}

Distinct evaluation methods lead to inconsistent score distributions and scopes amongst various sub-datasets, thus making it a challenging task to reasonably combine all scores. Rank standard normalization is a frequently used approach for score normalization:

\begin{equation}
\begin{aligned}
\mathbf{s}^{rank} &= \frac{\mathrm{rank}(\mathbf{s})}{\mathrm{len}(\mathbf{s})},\\
\mathbf{s}^{norm} &= \frac{\mathbf{s}^{rank} - \mu^{rank}}{\sigma^{rank}},
\end{aligned}
\end{equation}
where $\mathbf{s}$ and $\mathbf{s}^{norm}$ are the original and normalized score vectors of all models, $\mathrm{rank}()$ is the function to rank the scores, $\mathrm{len}()$ is the function to compute the length of $\mathbf{s}$, $\mu^{rank}$ and $\sigma^{rank}$ are the mean and standard derivation of $\mathbf{s}^{rank}$.

However, the above method eliminates the specific score differences between models, failing to accurately reflect the overall fundamental ability of LLMs. To address the issue, we propose a self-adaptive normalization method:
\begin{equation}
\begin{aligned}
\label{eq:norm}
s^{scale}_i &= \frac{s_i - \beta}{\alpha - \beta} * \gamma -\frac{\gamma}{2},\\
    s^{norm}_i &= \mathrm{Sigmoid}(s^{scale}_i) * 100,
\end{aligned}
\end{equation}
where $s_i$ and $s^{norm}_i$ are the original and normalized scores of the i-th model, $\alpha$ and $\beta$ are automatically calculated hyper-parameters based on the original scores, $\gamma$ is a hyper-parameter chosen by experiments. The proposed method aims to scale the original score of each LLM into an unify range in an self-adaptive way. The value for $\alpha$ and $\beta$ for each sub-dataset has been published on GitHub. Besides, we choose $\gamma$ = 2.5 based on experimental selection. More details of the normalization method are described in Appendix \ref{sec:norm_res_appendix}. 

After normalizing the scores for each sub-dataset, the final score of the model on F-Eval can be obtained by directly averaging.

\section{Experiments}

In this section, we conduct experiments to evaluate the performance on various LLMs on F-Eval using OpenCompass~\cite{2023opencompass}. Then, we pay our attention on two aspects: the evaluation methods' agreement with human judgements, and the distinction of the evaluation scores.

\paragraph{Settings}

When designing prompts, we directly provide the base model with texts that need to be continued or questions that need to be answered, without any additional instructions, which ensures that the evaluation of fundamental abilities is not limited by instruction-following abilities. For LLMs that default to a chat format, we add relevant instructions to the above prompts, such as "Please complete the text" or "Please answer the following question". In our experimental setup, three sub-datasets are evaluated in a few-shot setting. The ICL sub-dataset examine whether LLMs can induce information from in-context examples, while Commonsense Triple and Coreference use in-context examples to enable the model to learn for continuation, without explicit instructions. Apart from them, the remaining sub-datasets are all evaluated in a zero-shot setting. Specific prompts and settings for each sub-dataset are described in Appendix \ref{sec:appendix}.

\begin{figure*}[ht]
  \centering
  \includegraphics[width=0.85\textwidth]{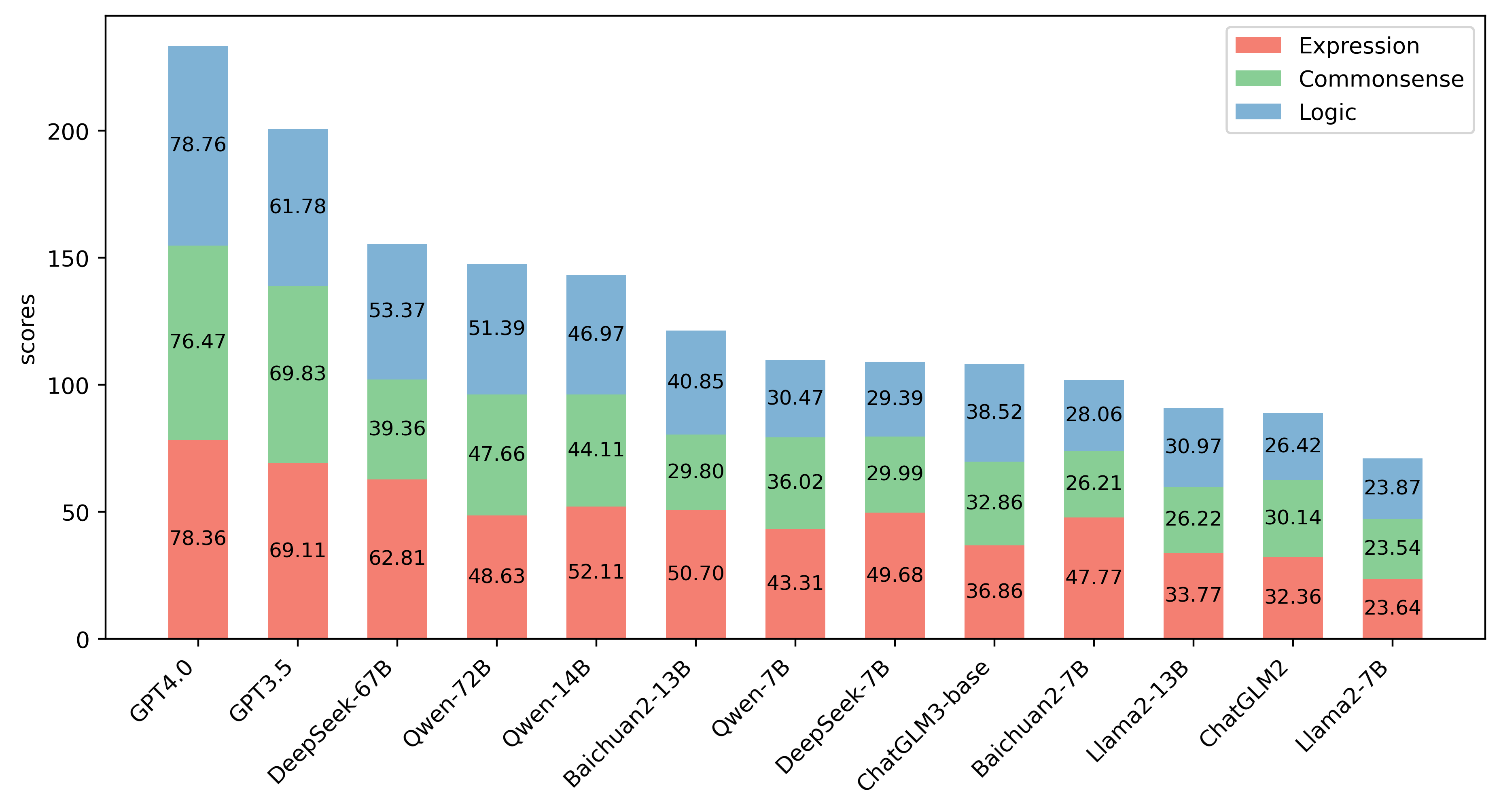}
  \caption{Main results of F-Eval across 13 LLMs.} \label{fig:results}
\end{figure*}

\paragraph{Models}
We evaluate 13 advanced LLMs from 6 model series in various sizes. For commercial models, we evaluate GPT series~\cite{chatgpt,openai2023gpt4}, while for open-source models, we select Llama2~\cite{touvron2023llama}, Baichuan2~\cite{baichuan2023baichuan2}, Qwen~\cite{qwen}, ChatGLM~\cite{du2022glm, zeng2022glm} and DeepSeek~\cite{DeepSeek-llm} series. Notably, we choose the base model of open-source models to better examine the fundamental abilities for the existence of alignment tax~\cite{DBLP:conf/nips/Ouyang0JAWMZASR22}. We introduce a detailed description of each LLM in Appendix \ref{sec:appendix_models}.

\subsection{Main Results}

The performance on F-Eval across LLMs is shown in Figure \ref{fig:results}. For clarity, we aggregate the results and report them in 3 dimensions in the figure. The detailed scores on 15 sub-datasets are listed in Appendix \ref{sec:results_appendix}.

As observed, GPT4.0 and GPT3.5 significantly outperform other models, achieving 78\% and 67\% correctness, respectively. However, with none of the open-source models achieving scores above 55\%, it is evident that they still encounter significant hurdles in their fundamental abilities. Among open-source models, DeepSeek and Qwen series exhibit superior performance compared to other series. Llama2 struggles to achieve only less than 30\% scores. Moreover, the results show that within each series, the performance of LLMs improves as the model size increases. When enlarging the model size from 7B to approximately 13B, the LLMs' performance improved by an average of 22\%. When further expanding the model size to around 70B, LLMs show much better performance, with 56\% improvement compared to the 7B model.

From the perspective of the 3 dimensions, each model demonstrates relatively good performance in expressive capabilities. Among them, DeepSeek-67B is particularly outstanding in expression dimension, closely approaching GPT3.5. Although open-source models obtain great expressive ability, they are still far behind the GPT series in dimensions of commonsense and logic. Qwen series slightly outperforms other LLMs with similar model size in applying commonsense and logic.

\subsection{Meta Evaluation}
\label{sec:correlation}

In order to evaluate the reliability of the evaluation methods of our benchmark, we utilize meta evaluation, which is performed in terms of Pearson correlation coefficient ($r$) ~\cite{mukaka2012guide} and Spearman correlation coefficient ($\rho$)~\cite{doi:https://doi.org/10.1002/0470011815.b2a15150} between human judgment and automated metrics. For all dimensions, we present sample-level correlations. Since the evaluation methods of objective tasks are unequivocally defined and undisputed, we only consider the subjective tasks. Given that manually annotating the entire dataset is costly and time-consuming, we sample around 300 instances as an approximation. 

\begin{table*}[htb]
\renewcommand\arraystretch{1.1}
  \centering
  \small
  \setlength{\tabcolsep}{3pt}
  \tabcolsep=0.3cm
  \begin{tabular}{@{}lcccccccc@{}}
    \toprule
        \multirow{2}{*}{Metrics}  & \multicolumn{2}{c}{Expression} & \multicolumn{2}{c}{Commonsense} & \multicolumn{2}{c}{Logic} & \multicolumn{2}{c}{Average}      \\
        \cmidrule(l{8pt}r{8pt}){2-3} \cmidrule(l{8pt}r{8pt}){4-5} \cmidrule(l{8pt}r{8pt}){6-7} \cmidrule(l{8pt}r{8pt}){8-9}
        & $r$ & $\rho$ & $r$ & $\rho$ & $r$ & $\rho$ & $r$ & $\rho$  \\
        \midrule
        BLEU & 0.224 & 0.197 & 0.306  & 0.361 & 0.011 & -0.016  & 0.180 & 0.181 \\
        BERTScore  & 0.632 & 0.623 & 0.618 & 0.638 & 0.469 & 0.255 & 0.573 & 0.505\\
        GPT4.0 & 0.585 & 0.414 & \textbf{0.918}  & \textbf{0.904} & 0.233 & 0.225 & 0.567 & 0.508 \\
        Auto-J & 0.584 & 0.489 & 0.895 & 0.818 & 0.473 & 0.449 & 0.651 & 0.585 \\
        \midrule
        F-Eval & &&&&&&&\\
        \quad w/ Rank standard  & 0.242 & 0.286 & 0.706 & 0.673 & 0.432 & 0.380 & 0.460 & 0.446\\
        \quad w/ Self-adaptive (ours) & \textbf{0.768} & \textbf{0.764} & \textbf{0.918}  & \textbf{0.904} & \textbf{0.706} & \textbf{0.557} & \textbf{0.797} & \textbf{0.742} \\
         
    \bottomrule
  \end{tabular}
  \caption{Comparison of Pearson ($r$) and Spearman ($\rho$) correlation coefficients, in expression, commonsense and logic dimensions. The upper block represents the baselines, while the lower block is our own method, where 'w/ Rank Standard' pertains to the statistical method of Rank Standard Normalization, and 'w/ Self-Adaptive' refers to the normalization method we design.}
  \label{tab:correlation}
\end{table*}

We evaluate our evaluation methods against two traditional metrics, \textbf{BLEU}~\cite{DBLP:conf/acl/PapineniRWZ02} and \textbf{BERTScore}~\cite{DBLP:conf/iclr/ZhangKWWA20}. Among them, BLEU is a ngram-based metric, while BERTScore is an embedding-based metric using BERT~\cite{DBLP:conf/naacl/DevlinCLT19}. Both of them compute the difference between reference texts and output texts for scoring. Considering the requirement of references, annotators are required to provide exemplars for reference-free sub-datasets. Apart from traditional methods, we also choose some top-performing evaluation methods based on LLMs, including GPT4.0~\cite{openai2023gpt4} and Auto-J~\cite{li2023generative}. The evaluation prompts of GPT4.0 are also adapted from \citet{zheng2023judging}. Auto-J is a generative judgement specifically trained for evaluation. The coefficient scores are shown in Table~\ref{tab:correlation}. "w/ Rank Standard" indicates results normalized by Rank Standard Normalization, while "w/ Self-Adaptive" uses the self-adaptive normalization methods designed by us. Notably, since subjective sub-datasets in commonsense dimensions are all reference-based, the coefficient scores of GPT4.0 and F-Eval is the same.

The results show that our evaluation methods consistently achieve higher correlation coefficient than other baselines in all dimensions. Specifically, the correlation of our methods far exceeds other baselines in the dimensions of expression and logic, proving that our newly designed evaluation method for reference-free tasks is superior to traditional and LLM-based scoring. While in the commonsense dimension, our results are on par with those of Auto-J. We observe that GPT4.0 exhibit slightly better performance on reference-based tasks. In our work, we utilize GPT4.0 considering its performance advantages and enhanced generalization abilities, while Auto-J can be employed as a budget-friendly substitution.

\subsection{Distinction}

\begin{figure}[h]
  \centering
  \includegraphics[width=\columnwidth]{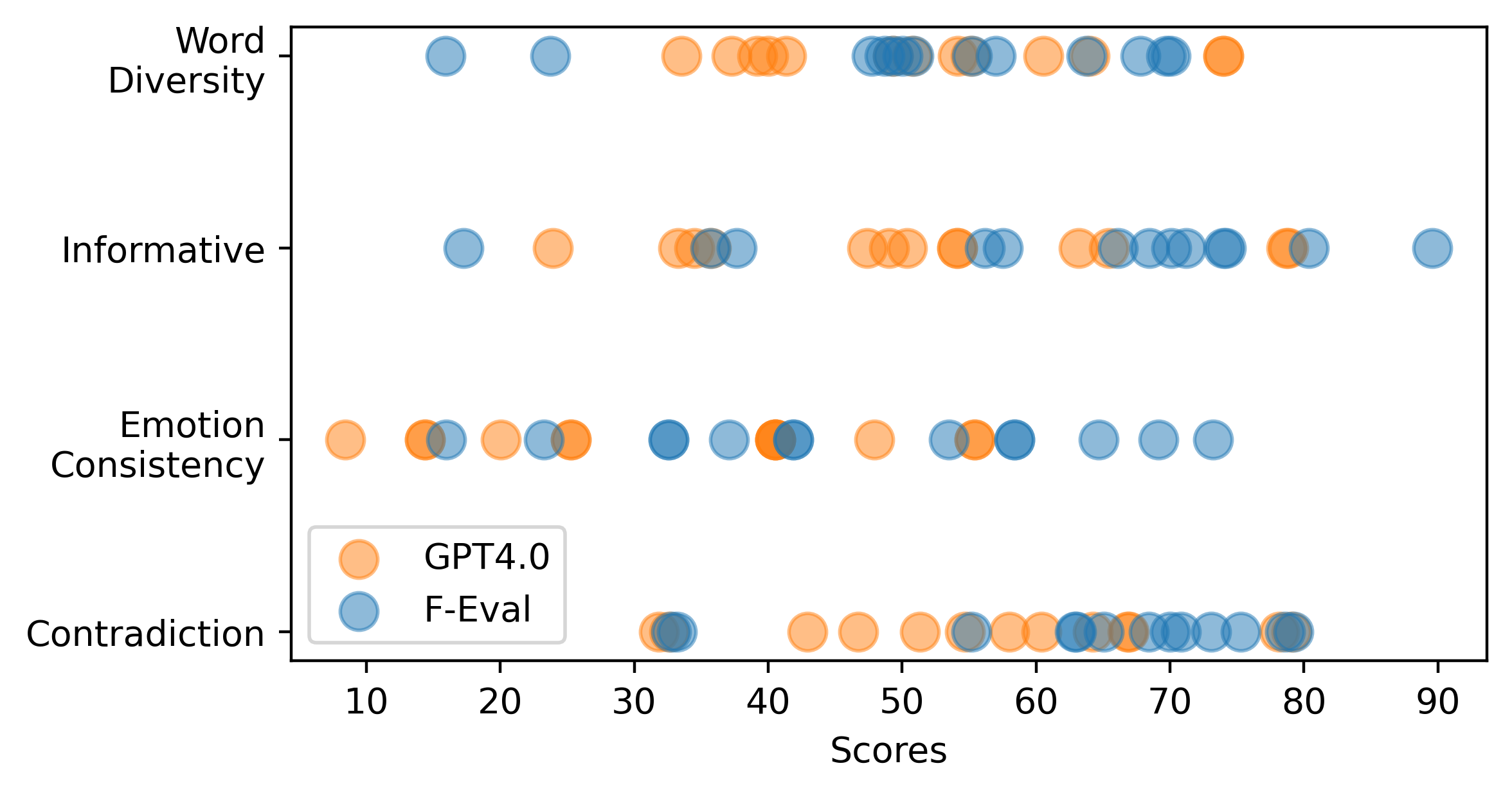}
  \caption{The distribution of the scores computed by GPT4.0 and F-Eval.} \label{fig:distinction}
\end{figure}

Researchers~\cite{zheng2023judging} have demonstrated that without providing references to API models, it becomes challenging for them to discern minor differences between responses, leading to a more concentrated distribution of scores and smaller distinction among models. To address this issue, we introduce new evaluation methods for reference-free subjective sub-datasets. To verify whether our evaluation method offers greater distinction, we visualize the distribution of scores within the reference-free sub-datasets (Figure~\ref{fig:distinction}) and also calculate the standard deviations and ranges (Table~\ref{tab:distinct}). Given the considerable expense associated with API Evaluations, we conduct experiments on select datasets as mentioned in Section~\ref{sec:correlation}.

\begin{table}[h]
\renewcommand\arraystretch{1.1}
  \centering
  \small
  \setlength{\tabcolsep}{3pt}
  \tabcolsep=0.1cm
  \begin{tabular}{@{}lcccc@{}}
    \toprule
         & \multicolumn{2}{c}{Standard deviation} & \multicolumn{2}{c}{Range} \\
         \cmidrule(l{4pt}r{4pt}){2-3} \cmidrule(l{4pt}r{4pt}){4-5}
         & GPT4.0 & F-Eval & GPT4.0 & F-Eval\\
        \midrule
        Word Diversity & 13.50 & \textbf{16.33} & 40.48 & \textbf{54.12}\\
        Informative & 17.04 & \textbf{21.21} & 54.90 & \textbf{75.73}\\
        Emotion Consistency & 15.90 & \textbf{17.98} & 46.95& \textbf{57.25}\\
        Contradiction & 15.16 & \textbf{15.17} & \textbf{47.20}& 46.47\\
          \bottomrule
  \end{tabular}
  \caption{The comparison of the standard deviation and range of scores between GPT4.0 and F-Eval on reference-free subjective sub-datasets.}
  \label{tab:distinct}
\end{table}

As we can observe from Figure~\ref{fig:distinction}, 
the scoring distribution obtained through F-Eval is notably more dispersed, unlike the GPT4.0 outcomes which tend to cluster around certain score ranges. This is corroborated by the standard deviation in Table~\ref{tab:distinct}. The standard deviations of our evaluation methods exceed those of GPT4.0 across all sub-datasets, signifying that our approach yields greater score variations in response to differences in model outputs. Moreover, as shown in Figure~\ref{fig:distinction}, unlike GPT4.0, our evaluation methods feature a broader scoring range, as reflected by the larger range in Table~\ref{tab:distinct}. Hence, it is evident that the scoring for reference-free subjective sub-datasets in F-Eval offers more distributed results compared to GPT4.0, thereby more accurately reflecting the differences between various LLMs.

\section{Discussion}

\begin{figure}[h]
  \centering
  \includegraphics[width=\columnwidth]{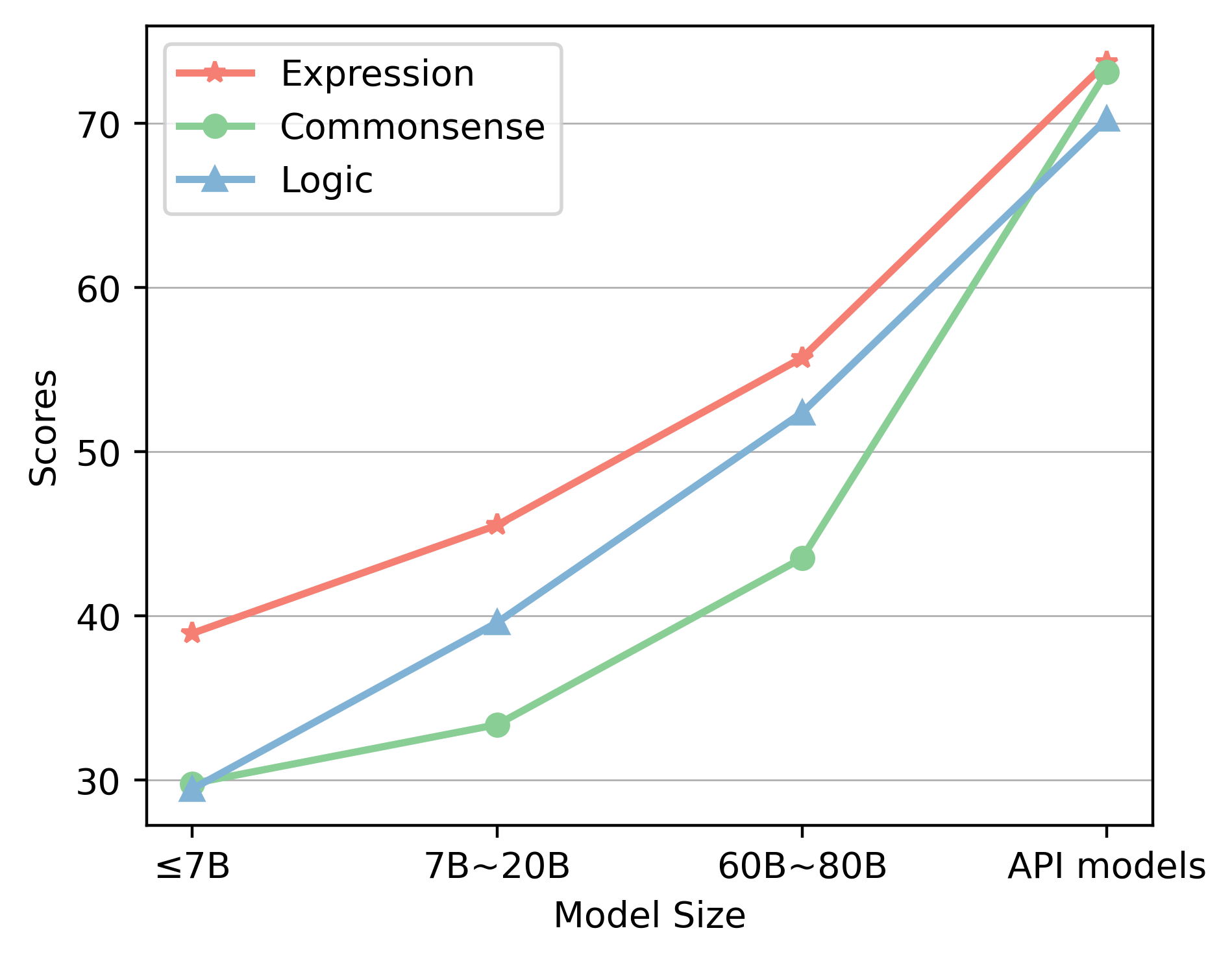}
  \caption{The impact of the model size in each dimensions.} \label{fig:model-size}
\end{figure}

\paragraph{The impact of the model size on performance across three dimensions.}
In order to investigate how the performance of LLMs improves with the increase in the model size, we categorize the selected open-source models into three levels: small-scale models with 7B or less, medium-scale models between 7B and 20B, and large-scale models ranging from 60B to 80B. We depict the score trends of open-source models across three model scales, as well as those of API models in Figure \ref{fig:model-size}. It is clear that the performance in each dimension increases with the enlargement of the model size. Specifically, we observe that increasing the model size from small scale to medium scale obviously enhances the ability of logic, while the performance of expression and commonsense only have a tiny improvement. With further enlargement, a substantial improvement is observed in all dimensions. Based on the observation above, we speculate that with parameters bigger than 80B, the model should exhibit better fundamental capabilities. However, the figure clearly shows that current open-source LLMs significantly lag behind API models in every dimension. Therefore, there is still a considerable journey ahead in our exploration of LLMs.

\paragraph{The ability imbalance of each LLM across three dimensions.}

As shown in Figure \ref{fig:results}, the overall performance of the LLMs is not completely consistent with its performance in each dimension. To further explore whether each model's abilities are balanced across 3 dimensions, we compare rankings of the overall results and those in each dimension in Table \ref{tab:imbalance}. The ranking proves that GPT4.0 and GPT3.5 consistently outperform other open-source models in every dimension. Llama2 series exhibits suboptimal performance in almost all dimensions, with only subtle improvement in logic ability. Additionally, we observe that DeepSeek and Baichuan2 series excel in expression, while show notable shortcomings in commonsense. Conversely, Qwen and ChatGLM series show better ability on commonsense and fail on expression. Notably, every LLM series demonstrates that larger-scale models exhibit obviously superior capabilities in logic compared to their smaller counterparts. The conclusion is also consistent with the observations seen in Figure~\ref{fig:model-size}.

\begin{table}[htb]
\renewcommand\arraystretch{1.1}
  \centering
  \small
  \setlength{\tabcolsep}{3pt}
  \tabcolsep=0.1cm
  \begin{tabular}{@{}lcccc@{}}
    \toprule
        Models & Overall & Expression & Commonsense & Logic      \\
        \midrule
        GPT4.0 & 1 & 1 & 1 & 1\\
        GPT3.5 & 2 & 2 & 2 & 2\\
        DeepSeek-67B & 3 & 3 & 5 (\textcolor{red}{$\downarrow$}) & 3\\
        Qwen-72B & 4 & 7 (\textcolor{red}{$\downarrow$}) & 3 (\textcolor{green}{$\uparrow$}) & 4\\
        Qwen-14B  & 5 & 4 (\textcolor{green}{$\uparrow$}) & 4 (\textcolor{green}{$\uparrow$}) & 5\\
        Baichuan2-13B & 6 & 5 (\textcolor{green}{$\uparrow$}) & 10 (\textcolor{red}{$\downarrow$})& 6  \\
        DeepSeek-7B  & 7 & 6 (\textcolor{green}{$\uparrow$}) & 9 (\textcolor{red}{$\downarrow$}) & 10 (\textcolor{red}{$\downarrow$}) \\
        Qwen-7B & 8 & 9 (\textcolor{red}{$\downarrow$}) & 6 (\textcolor{green}{$\uparrow$}) & 9 (\textcolor{red}{$\downarrow$})\\
        ChatGLM3-base & 9 & 10 (\textcolor{red}{$\downarrow$}) & 7  & 7 \\
        Baichuan2-7B & 10 & 8 (\textcolor{green}{$\uparrow$}) & 11 (\textcolor{red}{$\downarrow$})& 11 (\textcolor{red}{$\downarrow$})\\
        Llama2-13B & 11 & 11 & 11 & 8 (\textcolor{green}{$\uparrow$})\\
        ChatGLM2 & 12 & 12 & 8 (\textcolor{green}{$\uparrow$}) & 12 \\
        Llama2-7B & 13 & 13 & 13 & 13 \\
         
          \bottomrule
  \end{tabular}
  \caption{The ranking of the overall results and those in each dimension. If the ranking in the current dimension is higher than the overall ranking, it is indicated with \textcolor{green}{$\uparrow$}; conversely, \textcolor{red}{$\downarrow$} is used.}
  \label{tab:imbalance}
\end{table}

\paragraph{Self-adaptive normalization v.s. rank standard normalization}

As mentioned in Section \ref{sec:norm}, we design a self-adaptive normalization method to substitute the Rank Standard Normalization. To compare these two normalization methods, we also present the correlation coefficient when using Rank Standard Normalization in Table \ref{tab:correlation}, marked as "w/ Rank Standard". The results show that simply using rank to normalize scores results in significantly lower correlation compared to the self-adaptive method. This is due to the fact that using rankings obscures the detailed differences between models within each sub-dataset, leading to the overall scores that can not accurately reflect the actual capabilities of the models. Our self-adaptive normalization method dynamically adjusts the scaling of scores based on the distribution of results in every sub-datasets. In this way, the differences between models within each sub-dataset are proportionally scaled, ultimately providing an accurate reflection of LLMs' fundamental capabilities.

\section{Conclusion}

We introduce F-Eval, a bilingual evaluation benchmark that focuses on the fundamental abilities of large language models within 3 dimensions, covering both objective and subjective tasks. For reference-free subjective tasks, we design more distinctive evaluation methods as an alternative to API scoring. Additionally, we develope a new self-adaptive normalization method to accurately and effectively combine scores from different sub-datasets. Experiments have shown that F-Eval's correlation coefficients across 13 advanced LLMs surpass those of other evaluation baselines. We hope our benchmarks can empower researchers to better enhance the fundamental abilities of LLMs during every stage.

\section*{Ethics Statement}

In this section, we clarify the main ethical statements of F-Eval. When constructing the dataset in F-Eval, the online data in F-Eval is collected from a public social media platform or websites, on which people can share or obtain information freely. The datasets we use for adaptation are all public and free for academic purpose, which are under licenses like MIT and CC BY-NCND 4.0 licenses. There are totally 9 annotators participating in the annotation work, with 3 experts and 6 ordinary participants. All annotators agree that their efforts will be used to build F-Eval, and they are paid according to their workload and passing rate. Details of the annotation is described in Appendix~\ref{sec:anno}. To protect the security and privacy of the data, F-Eval will be published only for academic researchers. We plan to publicly release the data in F-Eval under the CC BY-NCND 4.0 license.

\section*{Limitation}

We propose a benchmark for evaluating the fundamental capabilities of LLMs, primarily focusing on expression, commonsense, and logic capabilities. The selection of these three dimensions is empirical and may not fully cover all the fundamental capabilities that LLMs need to possess. Future work could expand the evaluation based on linguistics. Moreover, from the correlation coefficients, it can be seen that the evaluation methods for logic capabilities have slightly lower consistency with human scoring than the average, suggesting that future work could specifically research and innovate on subjective evaluation methods related to the logic. Overall, the proposed F-Eval enhances the evaluation totem for LLMs, filling the gap in objective and subjective evaluation tasks for base models. In the future, researchers can use F-Eval to monitor the fundamental capabilities of LLMs during each stage of training.

\section*{Acknowledgements}

This work is supported by Shanghai Artificial Intelligence Laboratory (No. P23KN00621). Sincerely appreciate the thoughtful comments by the reviewers of our paper. Special thanks to Songyang Zhang, Qiuyinzhe Zhang and all the annotators, whose invaluable efforts have greatly contributed to this work.

\bibliography{anthology,custom}

\appendix

\section{Details of the Benchmark}
\label{sec:appendix}

In this section, we describe the process of data collection and metrics of each sub-dataset from three dimensions in details. We also provide an example for each sub-dataset, each of which includes a prompt, the output from the LLM, and a reference answer (if available). 
Detailed statistics about sub-datasets in F-Eval are presented in Table \ref{tab:data_statistics}.

\begin{table*}[!ht]
\renewcommand\arraystretch{1.2}
  \setlength{\tabcolsep}{5pt}
  \centering
  \small
  \begin{tabular}{llccll}
  \toprule
    \multirow{2}{*}{Dimension} & \multirow{2}{*}{Sub-dataset} & \multicolumn{2}{c}{\#Samples} & \multirow{2}{*}{Task Format } & \multirow{2}{*}{Settings} \\
    \cmidrule(l{4pt}r{4pt}){3-4}
    & & \#English & \#Chinese & & \\
    \midrule
    \multirow{4}{*}{Expression} & Word Diversity & 51 & 102 & reference-free subjective & zero-shot\\
     & Informative & 72 & 111 & reference-free subjective & zero-shot \\
    & Rule Following & 66 & 75 & open-ended objective & zero-shot \\
     & Emotion Consistency & 70 & 80 & reference-free subjective & zero-shot \\
    \midrule
    \multirow{5}{*}{Commonsense} & Commonsense Triple & 84 & 66 & reference-based subjective & few-shot (k=5)\\
     & CommonsenseQA & 74 & 76 & multi-choice objective & zero-shot \\
    & TextbookQA & 75 & 76 & reference-based subjective & zero-shot \\
     & Story & 75 & 75 & multi-choice objective & zero-shot \\
    & Instruction & 80 & 70 & reference-based subjective & zero-shot \\
    \midrule
    \multirow{6}{*}{Logic} &  ICL & 75& 75& open-ended objective &few-shot (k=0,4)\\
     & COT & 80& 80& open-ended objective & zero-shot\\
    & Fallacy Attack & 52& 52& reference-based subjective & zero-shot \\
     & Contradiction & 75& 75& reference-free subjective & zero-shot\\
    & Coreference & 57& 58& open-ended objective & few-shot (k=4)\\
    & Anomaly Detection & 79& 75& multi-choice objective & zero-shot\\
  \bottomrule
  \end{tabular}
  \caption{The statistics and settings of each sub-dataset in F-Eval. } \label{tab:data_statistics}
\end{table*}

\subsection{Expression}

\subsubsection{Word Diversity}

\paragraph{Introduction}
Word choice is a fundamental linguistic capability. We evaluate a model's capability for word choice by examining whether it can utilize complex, advanced vocabulary, idioms, and proverbs. Test items will include an array of text types, such as prose, poetry, and classical literature, all characterized by their rich use of language. When extending or elaborating upon these texts, the model is expected to maintain a commensurate standard of eloquence in its word choice.
\paragraph{Data Collection}
We curate our test cases from a collection of online prose, poetry, news articles, and classical literature works that are published post-June 2023. Each test case is deliberately truncated and has undergone validation by human experts to confirm that the narratives are sufficiently open-ended. These steps guarantee that models are afforded ample creative latitude for text completion.
\paragraph{Evaluation Method}
Inspired by the approaches to assessing human writing abilities in school, we treat the usage of advanced vocabulary as a criterion for measuring word diversity. The advanced vocabulary is determined by textbooks and guidelines of human examinations. Since the vocabulary includes both phrases and single words, we will match the generated content with phrases first during the evaluation. If it fails to match any phrase, we evaluate the single words. 

\begin{table}[htb]
\small
    \centering
    \colorbox{gray!8}{
    \begin{tabular}{@{}p{7.3cm}}
    ======== \textsc{Prompt for Word Diversity} ========\\\\
    \textbf{Prompt:} \begin{CJK*}{UTF8}{gbsn}夜来策杖立中庭，仰望天边无数星。风卷白云千匹马，峰淹浩海一浮萍。黄金散去皆成梦，绿水横流岂有形？\end{CJK*}\\\\

    \textbf{Output: }\begin{CJK*}{UTF8}{gbsn}欲问人间何所似，只应长作梦中人。这首诗是宋代诗人王安石的《夜来》。王安石，字介甫，号半山，抚州临川人。北宋著名政治家、思想家、文学家、改革家，唐宋八大家之一。王安石在宋神宗熙宁年间任宰相，主持变法，史称“王安石变法”。\end{CJK*}\\\\

    \end{tabular}}
\end{table}

\subsubsection{Informative}

\paragraph{Introduction}
Generating fluent text is the most fundamental capability of language models, but simply examining the fluency of text is no longer sufficient to assess performance differences between large models. Models are expected to generate text that is not only fluent but also provides meaningful new content. Paraphrasing the previous context can generate fluent but not informative content. Therefore, we propose the informativeness of generated content as the metric for basic language quality. 

\paragraph{Data Collection}
We curate our test cases from journal articles, novels, and argumentative analysis essays published post-June 2023. Each test case is deliberately truncated and has undergone validation by human experts to confirm that the narratives are sufficiently open-ended. We also verify the succeeding context's informativeness is consistent with the previous using our metric. This step is necessary because the summary paragraph often has a very different informativeness compared to normal paragraphs. 
\paragraph{Evaluation Method}
We utilized a proxy LLM to evaluate the probabilities of the provided input prompt, $\mathbf{C}$, and the corresponding model-generated response, $\mathbf{X}$. The input prompt was hand-crafted to be open-ended, affording the model has enough flexibility for generating a completion. The expected model behavior is to produce output that maintains a consistent level of informativeness as the input, which is measured by the following metric:
\begin{equation}
\begin{aligned}
    &\mathrm{Info}(\mathbf{X}, \mathbf{C}) \\
    &= \lVert \frac{1}{|\mathbf{X}|} \log P(\mathbf{X} | \mathbf{C}) - \frac{1}{|\mathbf{C}|}\log P(\mathbf{C}) \rVert_1 .
\end{aligned}  
\end{equation}
In scenarios where $\frac{1}{|\mathbf{X}|} \log P(\mathbf{X} | \mathbf{C})$ is much smaller than $\frac{1}{|\mathbf{C}|}\log P(\mathbf{C})$, it suggests that the model's response lacks informativeness, such as paraphrasing, summarizing, or repeating the input. Conversely, when the model's responses are much more informative than the prompt, it may imply that the model has changed the topic and introduced unrelated content. The experiments on choosing proxy LLM is detailed in Appendix \ref{sec:informative}.

\begin{table}[htb]
\small
    \centering
    \colorbox{gray!8}{
    \begin{tabular}{@{}p{7.3cm}}
    ========== \textsc{Prompt for Informative} ==========\\\\
    \textbf{Prompt:} In this way, adults ought to be cognizant and appreciate the tremendous difficulty and great pain that some children experience when attempting to explain something that is very sensitive in nature to them.Listening can facilitate the acknowledgement of children and lead adults to the realization that children are frequently eager to share their testimony.They will only do so if adults are genuinely interested in them.\\\

    \textbf{Output: } In the end, the author concludes that adults should be aware of the importance of listening to children and should be willing to listen to them.\\
                    The author’s purpose in writing this passage is to persuade adults to listen to children.\\
                    The author’s attitude toward the topic is positive.\\
                    The author’s tone is persuasive.\\
                    The author’s purpose is to persuade adults to listen to children.\\\\

    \end{tabular}}
\end{table}

\subsubsection{Rule Following}

\paragraph{Introduction}
Rule following is a suite of ten sub-datasets designed to assess the model's ability to discern and adhere to patterns presented in input examples. These tasks instruct the model to follow specific patterns like pronunciation, radicals, and the positioning of words within phrases and sentences. We employ these sub-datasets to evaluate the model's proficiency in recognizing these fundamental linguistic components, which we consider essential underpinnings for advanced writing skills.
\paragraph{Data Collection}
We curated the rule following questions by human experts with the guidance of dictionaries. Since the questions are hand-crafted, we believe they are purely unseen text for all LLMs. 
\paragraph{Evaluation Method}
We evaluate the test cases according to their specific pattern by rule-based classifiers. Each test case will be assigned a binary score, representing following the pattern or not.  

\begin{table}[htb]
\small
    \centering
    \colorbox{gray!8}{
    \begin{tabular}{@{}p{7.3cm}}
    ======== \textsc{Prompt for Rule Following} ========\\\\
    \textbf{Prompt:} last chance,last minute,last name,last laugh,last resort\\\\

    \textbf{Output: }last word,last straw,last minute\\\\

    \end{tabular}}
\end{table}

\subsubsection{Emotion Consistency}

\paragraph{Introduction}

Given a story that is overall positive or negative, the LLM is asked to continue the story, examine whether it can maintain a consistent emotional tone in the output text. 

\paragraph{Data Collection}

To acquire data with distinct emotional tendencies, we utilize the sentiment intensities of the story dataset proposed by~\citet{DBLP:conf/acl/LuoDYLCSS19}, which is adapted from ROCStories. To obtain Chinese emotional data, we prompt ChatGPT to rewrite the corresponding English examples in Chinese. Besides, we follow \citet{DBLP:journals/corr/abs-2305-07759} to leverage ChatGPT for story generation. Each story is generated by conditioning on three randomly picked keywords and an emotion flag that indicates the story has a happy ending or a bad ending. 

\paragraph{Evaluation Method}

We use Assistant-Model Evaluation for Emotion Consistency. During the evaluation, the input and the generated output are first segmented into sentences. Then, we use a finetuned multi-lingual emotion classification model\footnote{We use the HuggingFace model in url{https://huggingface.co/nanaaaa/emotion\_chinese\_english.}} to determine the positive or negative sentiment of each sentence. The proportion of positive sentiments is calculated. Afterward, if the proportion of positive sentiments in the input and output exceeds a certain threshold (0.2), it is judged as inconsistent.

\begin{table}[htb]
\small
    \centering
    \colorbox{gray!8}{
    \begin{tabular}{@{}p{7.3cm}}
    ====== \textsc{Prompt for Emotion Consistency} ======\\\\
    \textbf{Prompt:} jennifer wished to purchase her mom and wonderful birthday gift . she scoured the house for all the money she could find . in the end , she could only scrounge up a meager forty seven cents . she decided this wasn ' t enough and opted to color a picture . her mother loved the picture and jennifer decided it was the best gift .\\\\

    \textbf{Output: }Jennifer's mom was so happy with the gift that she decided to give her daughter a special treat . she took Jennifer to the mall and let her choose anything she wanted . Jennifer was thrilled and couldn 't wait to pick out her favorite things . She spent hours at the mall with her mom , trying on clothes and eating treats .\\After a long day of shopping , Jennifer and her mom returned home . They were both tired but happy . Jennifer was so happy with her\\\\

    \end{tabular}}
\end{table}

\subsection{Commonsense}

\subsubsection{Commonsense Triple}

\paragraph{Introduction}

By summarizing the information of triples in the commonsense knowledge graph, the LLM, when given a head entity, can enumerate tail entities that fit a specific relation. We then compare the predicted entities with tail entities in the knowledge graph to determine the LLM's grasp of simple triple-based commonsense knowledge.

\paragraph{Data Collection}

We collect knowledge graph information from ConceptNet~\cite{DBLP:conf/aaai/SpeerCH17} and extract all the triples therein. Subsequently, we manually select triples with meaningful relations. Finally, we transform the data into a format consisting of all tail entities under a fixed head entity and relation. When constructing the prompt, we use 5-shot in-context learning (ICL) prompt. The example below only show one-shot for short.

\paragraph{Evaluation Method}

We use API Evaluation for Commonsense Triple sub-dataset. The expected results being structured information, but the output formats of different LLMs are highly diversified and difficult to match with a unified rule for predicted entities. Therefore, we directly assign scores using the LLM's output and the gold entity. The specific evaluation methods are depicted in Figure \ref{fig:api_eval}. Among them, "Answer Type" is Factual Enumeration Question, "Evaluation Dimension" has Factuality, User Satisfaction, Richness and Completeness.

\begin{table}[htb]
\small
    \centering
    \colorbox{gray!8}{
    \begin{tabular}{@{}p{7.3cm}}
    ====== \textsc{Prompt for Commonsense Triple} ======\\\\
    \textbf{Prompt:} Entity: cow. Relation: locate at. Words that can form a corresponding relation with the entity: ['middle\_of\_eating\_grass', 'indiana', 'computer\_commercial', 'bard', 'outside\_in\_pasture', 'red\_barn', 'fiueld', 'america', 'outdoors', 'herd', 'nebraska', 'nursery\_rhyme']\\
Entity: cat. Relation: desire. Words that can form a corresponding relation with the entity:\\\\

    \textbf{Output: }['sleep', 'food', 'attention', 'love', 'cuddle', 'play', 'affection', 'nap', 'cuddling', 'affectionate', 'pet']\\\\

    \textbf{Gold:} [
            "milk\_to\_drink",
            "eat",
            "food",
            "meow",
            "petted"
        ]

    \end{tabular}}
\end{table}

\subsubsection{CommonsenseQA}

\paragraph{Introduction}

By embedding commonsense information from the knowledge graph into specific scenarios and transforming it into multiple-choice questions, we enable the model to choose the most suitable option from multiple choices. This tests the model's ability to grasp and discern commonsense knowledge.

\paragraph{Data Collection}

This section of data is primarily adapted from CommonsenseQA~\cite{DBLP:conf/naacl/TalmorHLB19}. To obtain Chinese data, we first translate the questions into Chinese using GPT3.5. Then, we process all the questions, both in Chinese and English, with InstructGPT\footnote{We use text-davinci-003 version for InstructGPT}, keeping only those that answer correctly to ensure the questions are not too difficult. Finally, we reshuffle the order of the options in the remaining questions to prevent LLMs from memorizing past answers.

\paragraph{Evaluation Method}

We use Probability Evaluation for CommonsenseQA. When evaluating, we append each answer after the question, and calculate their perplexity (PPL). Then we choose the option with the lowest PPL as prediction to obtain the accuracy. Notably, API models should directly output the option.

\begin{table}[htb]
\small
    \centering
    \colorbox{gray!8}{
    \begin{tabular}{@{}p{7.3cm}}
    ======== \textsc{Prompt for CommonsenseQA} ========\\\\
    \textbf{Prompt A:} Where are people likely to stand at an intersection?\\Answer: cars\\
    \textbf{Prompt B:} Where are people likely to stand at an intersection?\\Answer: city street\\
    \textbf{Prompt C:} Where are people likely to stand at an intersection?\\Answer: street corner\\
    \textbf{Prompt D:} Where are people likely to stand at an intersection?\\Answer: fork in road\\
    \textbf{Prompt E:} Where are people likely to stand at an intersection?\\Answer: at a red light\\\\

    \textbf{PPL A:} 6.53\\
    \textbf{PPL B:} 6.56\\
    \textbf{PPL C:} 6.02\\
    \textbf{PPL D:} 6.71\\
    \textbf{PPL E:} 5.77\\\\

    \textbf{Gold:} C

    \end{tabular}}
\end{table}

\subsubsection{TextbookQA}

\paragraph{Introduction}

Given questions based on commonsense knowledge appearing in elementary school textbooks, the LLM is tasked with answering. This assesses the model's grasp of knowledge-based commonsense from various subjects. 

\paragraph{Data Collection}

We collect original K12 Chinese data. Firstly, we clean it by removing pinyin, formulas, tables, images, and other distracting information to obtain pure text data. Then, we segment each data entry, mainly by chapters, and further divided every five paragraphs if the length is still long after the initial division. The segmented text is used as a prompt input for GPT3.5 to generate a commonsense question related to the text. We then use InstructGPT for screening, retaining only answerable questions. Through this process, we obtain the final Chinese version of the TextbookQA. For the English version, as we do not find suitable K12 English data, we randomly select three Chinese questions, prompting GPT3.5 mimic generating English questions and reference textbooks. Finally, we manually screen the generated English questions to obtain the final English TextbookQA.

\paragraph{Evaluation Method} 

Since the answers to the questions are included in the textbook, we can't directly use rules to judge whether the answers are correct. Therefore, we chose the API Evaluation method, allowing GPT4.0 to score the model's output based on the textbook. The scoring prompt is shown in Figure \ref{fig:api_eval}, where "Answer Type" is Factual and Explanatory Question, "Evaluation Dimension" has Factuality, User Satisfaction, Clarity and Completeness.

\begin{table}[htb]
\small
    \centering
    \colorbox{gray!8}{
    \begin{tabular}{@{}p{7.3cm}}
    ========= \textsc{Prompt for TextbookQA} =========\\\\
    \textbf{Prompt:} Question: What is the definition of hectare?\\Answer:\\\\

    \textbf{Output: }unit of area\\\\

    \textbf{Textbook:} To measure land area, we can use 'hectare' as a unit. The 'Bird's Nest' is really magnificent! Its area is about 20 hectares. The area of a square with a side length of 100 meters is 1 hectare. The area enclosed by a 400-meter running track is approximately 1 hectare.

    \end{tabular}}
\end{table}

\subsubsection{Story}

\paragraph{Introduction}

Given the first half of a story and two possible endings, the LLM is tasked with choosing the correct ending that aligns with commonsense. This primarily examine the model's ability to judge whether the development of a story in a specific context is reasonable.

\paragraph{Data Collection}

We adapt the ROCStories dataset~\cite{DBLP:journals/corr/MostafazadehCHP16}, which is divided into a train set, a validation set, and a test set. Each instance contains a 4-sentence story, where the train set has only one correct ending, the validation set has one correct and one incorrect ending, and the test set has two endings without correctness labels. To standardize it into a usable format, we use GPT3.5 to generate incorrect endings for the stories in the train set and select correct endings for the stories in the test set. Finally, we merge the three sets and randomly selecte stories for our Story sub-dataset.

\paragraph{Evaluation Method}

We use Probability Evaluation for Story. When evaluating, we append each ending after the story, and calculate their perplexity (PPL). Then we choose the ending with the lowest PPL as prediction to obtain the accuracy. Notably, API models should directly output the option.

\begin{table}[htb]
\small
    \centering
    \colorbox{gray!8}{
    \begin{tabular}{@{}p{7.3cm}}
    ============ \textsc{Prompt for Story} ============\\\\
    \textbf{Prompt A:} Megan and I walked home from school near a busy street. I saw three blue cars pass by in a row. We decided to count the blue cars. Megan and I sat on the curb all afternoon. After counting the cars, we went back home.\\
    \textbf{Prompt B:} Megan and I walked home from school near a busy street. I saw three blue cars pass by in a row. We decided to count the blue cars. Megan and I sat on the curb all afternoon. We then sat down and began to count the cars,\\\\

    \textbf{PPL A:} 2.85\\
    \textbf{PPL B:} 3.00\\\\

    \textbf{Gold:} A
    \end{tabular}}
\end{table}

\subsubsection{Instruction}

\paragraph{Introduction}

Instruction is designed to assess the LLM's ability to understand and follow simple instructions, which maintain commonsense knowledge.

\paragraph{Data Collection}

The English data is adapted from Alpaca~\cite{alpaca} and the Chinese data is adapted from Alpaca-zh~\cite{peng2023instruction}. Similar to the data filtering process for TextbookQA, the Instruction data also initially undergoes a screening using InstructGPT to filter out instructions that can be correctly executed, followed by a manual secondary screening to obtain the final dataset.

\paragraph{Evaluation Method}

The evaluation on Instruction are the same as that of TextbookQA. The scoring prompt is shown in Figure \ref{fig:api_eval}, where "Answer Type" is Factual and Explanatory Question, "Evaluation Dimension" has Factuality, User Satisfaction, Clarity and Completeness.

\begin{table}[htb]
\small
    \centering
    \colorbox{gray!8}{
    \begin{tabular}{@{}p{7.3cm}}
    ========== \textsc{Prompt for Instruction} ==========\\\\
    \textbf{Prompt:} Rewrite the sentence with more descriptive words. The game is fun.\\\\

    \textbf{Output: }The game is enjoyable.\\\\

    \textbf{Gold:} The game is incredibly engaging and enjoyable.

    \end{tabular}}
\end{table}

\subsection{Logic}

\subsubsection{ICL}
\paragraph{Introduction} In our evaluation of previous models using ICL, we focus on whether the model can deliver results in the given example format. Our goal is to determine how much the model's performance improves with the increase in the number of examples. If the performance improves quickly, it suggests that the model has strong generalization capabilities and induction abilities.

\paragraph{Data Collection} Our primary dataset is the NaturalQuestions~\cite{47761} (NQ) dataset. We manually filter out specific samples related to time, and then translate them to form a Chinese version.

\paragraph{Evaluation Method} It belongs to Rule-based Evaluation. We use the exact match method to compare the generated results and gold answers. We have considered two shot categories in our experiment, including 0-shot and 4-shot. We consider both the absolute scores and the incremental scores using 4-shot when computing the final results:
\begin{equation}
\begin{aligned}
\mathrm{Result} &= \frac{{x4 - x0}}{3} + \frac{{2 \cdot x4}}{3},
\end{aligned}
\end{equation}
where $x4$ and $x0$ are the scores of the 4-shot and 0-shot models respectively. Indeed, since LLMs with poorer capabilities have lower initial scores ($x0$), the metric $x4-x0$ is more favorable to them. Therefore, to preserve the true capability of the LLMs, we consider slightly increasing the proportion of $x4$. Experiments show the Spearman correlation scores of ICL under different weights are essentially unchanged under different weights. We ultimately chose the weights of $x4$ as 2/3, which is alternative.

\begin{table}[htb]
\small
    \centering
    \colorbox{gray!8}{
    \begin{tabular}{@{}p{7.3cm}}
    ============= \textsc{Prompt for ICL} ==============\\\\
    \textbf{Prompt:} Answer the question, your answer should be as simple as possible, start your answer with the prompt "The answer is ". \\Q: Who sings does he love me with reba?\\A: The answer is Linda Davis.\\Answer the question, your answer should be as simple as possible, start your answer with the prompt "The answer is ". \\Q: who got the first Noble Prize in physics?\\A: \\\\

    \textbf{Output:} The answer is Wilhelm Conrad Röntgen. \\\\

    \textbf{Gold:} Wilhelm Conrad Röntgen \\\\

    \end{tabular}}
\end{table}

\subsubsection{COT}
\paragraph{Introduction} In the methodologies that previous models use with COT, the focus is solely on how COT deduces the correct answer. We introduce a slight modification, aiming to assess whether LLMs have the reasoning ability to comprehend the chain-of-thought process and make correct prediction. Consequently, we present the complete version of COT without the answer initially and then evaluate whether the model can infer the correct answer.

\paragraph{Data Collection} Our primary dataset is the GSM8K~\cite{cobbe2021gsm8k} dataset, from which we extract 160 questions for GPT4.0 to answer. Subsequently, we select questions with correct answers from the generated responses to establish the initial question bank. The answer typically embeds in the penultimate sentence (the last sentence often repeats the answer, resembling a student adding a concluding sentence after completing a question: "Therefore, the answer is..."). We remove sentences containing the answer to construct the final dataset.

\paragraph{Evaluation Method} It belongs to Rule-based Evaluation. Since the intermediate process is provided, we expect the model to generate the answer in the first sentence of the subsequent output. The model's response is considered correct if the answer is present in the first sentence.

\begin{table}[htb]
\small
    \centering
    \colorbox{gray!8}{
    \begin{tabular}{@{}p{7.3cm}}
    ============= \textsc{Prompt for COT} ==============\\\\
    \textbf{Prompt:} Janet's ducks lay 16 eggs per day. She eats three for breakfast every morning and bakes muffins for her friends every day with four. She sells the remainder at the farmers' market daily for \$2 per fresh duck egg. How much in dollars does she make every day at the farmers' market? \\

    Every day, Janet uses 3 eggs for breakfast and 4 for muffins, a total of 3 eggs + 4 eggs = <<3+4=7>>7 eggs. This means she has 16 total eggs - 7 used eggs = <<16-7=9>>9 remaining eggs to sell. \\\\

    \textbf{Output:} Janet can earn \$18 every day. \\\\

    \textbf{Gold:} 18 \\\\

    \end{tabular}}
\end{table}

\subsubsection{Fallacy Attack}
\paragraph{Introduction} Given two physical objects that have similarities in some respects but not in the larger direction, the task is to test whether the model is misled by these seemingly similar but fundamentally wrong reasoning steps, resulting in statements that violate commonsense.

\paragraph{Data Collection} We manually construct some mutually connected yet contradictory examples, and then use GPT to construct similar data.

\paragraph{Evaluation Method} This belongs to API Evaluation. The generated sentences are input into GPT4.0 for scoring, allowing it to rate the model's responses from 0 to 10. The key points are two-fold, including whether the generated sentence violates commonsense and whether the model's output and the reference answer are semantically consistent.

\begin{table}[htb]
\small
    \centering
    \colorbox{gray!8}{
    \begin{tabular}{@{}p{7.3cm}}
    ======= \textsc{Prompt for Fallacy Attack} =======\\\\
    \textbf{Prompt:} Humans can speak, babies are humans, so \\\\

    \textbf{Output:} Babies can also speak. \\\\
    \end{tabular}}
\end{table}

\subsubsection{Contradiction}

\paragraph{Introduction} Given a biographical narrative that contains a wealth of character information, the task is to let the model generate a paragraph of text and then examine whether the output has contradiction.

\paragraph{Data Collection} The dataset is sourced from the Wiki-bio~\cite{DBLP:journals/corr/LebretGA16} dataset by HuggingFace. We first filter out samples from this dataset that are longer than 512. Then, we translate these samples into Chinese. This process forms the dataset format we require.

\paragraph{Evaluation Method} It belongs to Assistant-Tool Evaluation. Initially, the input and output are amalgamated, following which each sentence is sequentially paired with its preceding textual context. Then, we use the off-shelf Natural Language Inference (NLI) models~\cite{laurer_van} to determine whether each sequence pair is contradiction. 

\begin{table}[htb]
\small
    \centering
    \colorbox{gray!8}{
    \begin{tabular}{@{}p{7.3cm}}
    ========= \textsc{Prompt for Coreference} ==========\\\\
    \textbf{Prompt:} Roger Ducret (April 2, 1888 - January 8, 1962) was a French fencer who participated in the 1920, 1924, and 1928 Olympics. At the 1924 Summer Olympics, he participated in five of the six events, winning a gold or silver medal in each event, and won individual medals in three fencing competitions: epee, foil, and sabre. In history, only one fencer did better--Nedo Nadi won five gold medals at the 1920 Summer Olympics. During the First World War, Ducret was a prisoner of war. After retiring, he worked as a journalist for the Figaro, Echoes of Sport, and other newspapers. \\\\

    \textbf{Output:} Ducret was born in Paris, France, and grew up in the Saint-Germain-des-Prés district of Paris. He began his fencing career in 1908 and won his first Olympic medal, a bronze, in 1912. At the 1920 Olympics, he won two gold medals and one silver medal. At the 1924 Olympics, he won three gold medals and one silver medal. Ducret won two gold medals, one silver medal, and one bronze medal at the 1920 Olympics. \\\\
    \end{tabular}}
\end{table}

\subsubsection{Coreference}
\paragraph{Introduction} This dataset mainly aims to see whether the model can clearly distinguish whether the different pronouns correspond to the correct characters when multiple characters appear.

\paragraph{Data Collection} The dataset is transformed from the winograd\_wsc~\cite{levesque2012winograd} dataset on HuggingFace. The transformation method is to select the last short sentence that contains a specific pronoun, and then add after this short sentence: He/She/It refers to...

\paragraph{Evaluation Method} 

During the evaluation, 4 shots will be provided in the prompt. These shots will follow the pronoun with brackets and the referred names, which the model will learn from. We will truncate a sentence after the pronoun and the left bracket. The model should generate a name and a right bracket based on the context. In the end, it only needs to judge whether the generated name matches the answer.

\begin{table}[htb]
\small
    \centering
    \colorbox{gray!8}{
    \begin{tabular}{@{}p{7.3cm}}
    ======== \textsc{Prompt for Contradiction} =========\\\\
    \textbf{Prompt:} The trophy does not fit in the brown suitcase because it is too large. It refers to (the trophy). Paul tried to call George by phone, but he was not there. He refers to (George). The lawyer asked the witness a question, but he (the witness) did not want to answer. He refers to (the witness). Anna performed much worse in the exam than her good friend Lucy, because she studied too hard. She refers to (Lucy). Peter is jealous of Martin, even though he is very successful. He refers to (\\\\

    \textbf{Answer:} Martin). \\\\
    
    \textbf{Gold:} Peter \\\\
    \end{tabular}}
\end{table}

\subsubsection{Anomaly Detection}
\paragraph{Introduction} This dataset aims to verify a model function similar to Coreference sub-dataset. However, while Coreference asks the model to generate a specific noun, anomaly detection requires the model to discern the perplexity of correct sentences and sentences with pronoun errors, thus selecting the correct sentence.

\paragraph{Data Collection} The dataset is transformed from the winograd\_wsc~\cite{levesque2012winograd} dataset on HuggingFace, selecting examples that meet the requirements and translating them into Chinese.

\paragraph{Evaluation Method} We use Probability Evaluation for Anomaly Detection. When evaluating, we append each coreference option, and calculate their perplexity (PPL). Then we choose the option with the lowest PPL as prediction to obtain the accuracy. Notably, API models should directly output the option.

\begin{table}[htb]
\small
    \centering
    \colorbox{gray!8}{
    \begin{tabular}{@{}p{7.3cm}}
    ====== \textsc{Prompt for Anomaly Detection} ======\\\\
    \textbf{Prompt A:} The city councilmen refused the demonstrators a permit because they feared violence.'they'refer to The city councilmen. \\\\
    
    \textbf{Prompt B:} The city councilmen refused the demonstrators a permit because they feared violence.'they'refer to The demonstrators. \\\\
    
    \textbf{PPL A:} 3.51\\
    \textbf{PPL B:} 3.68\\\\
    
    \textbf{Gold:} A \\\\
    \end{tabular}}
\end{table}

\section{Detailed Evaluation Methods}

The detailed evaluation methods of all sub-datasets are described in Appendix \ref{sec:appendix}. In this section, we show the choice of the proxy LLMs in Informative dataset and prompts using for API Evaluation. Besides, we detail the normalization methods for the overall scores.

\subsection{Informative}
\label{sec:informative}

We use a proxy LLM as the assistant tool when evaluating Informative dataset. In principle, the selection only requires the use of open-source LLMs with good language capabilities, and the model size does not need to be particularly large. To choose a more suitable proxy LLM, we conduct experiments on correlation coefficients using DeepSeek-7B, Baichuan2-7B, and ChatGLM3-base, with the results listed in Table \ref{tab:informative}. From the results, it can be seen that the results among different proxy LLMs do not vary significantly, indicating that the evaluation method of Informative sub-dataset has good robustness. Since the score of DeepSeek-7B is slightly higher, we opt to use it in F-Eval.

\begin{table}[htb]
\renewcommand\arraystretch{1.1}
  \centering
  \small
  \setlength{\tabcolsep}{3pt}
  \tabcolsep=0.3cm
  \begin{tabular}{@{}lcc@{}}
    \toprule
        \multirow{2}{*}{proxy LLM}  & \multicolumn{2}{c}{Expression}     \\
        \cmidrule(l{8pt}r{8pt}){2-3} 
        & $r$ & $\rho$ \\
        \midrule
         DeepSeek-7B  & \textbf{0.852} & \textbf{0.714} \\
        Baichuan2-7B  & 0.846 & 0.632\\
        ChatGLM3-base  & 0.807 & \textbf{0.714}  \\
         
    \bottomrule
  \end{tabular}
  \caption{Comparison of Pearson ($r$) and Spearman ($\rho$) correlation coefficients in Informative dataset when using different proxy LLMs.}
  \label{tab:informative}
\end{table}

\definecolor{my_orange}{RGB}{234,125,49}
\definecolor{my_blue}{RGB}{143,170,220}
\definecolor{my_yellow}{RGB}{255,204,0}
\begin{figure*}[htb]
\centering
\includegraphics[width=0.95\textwidth]{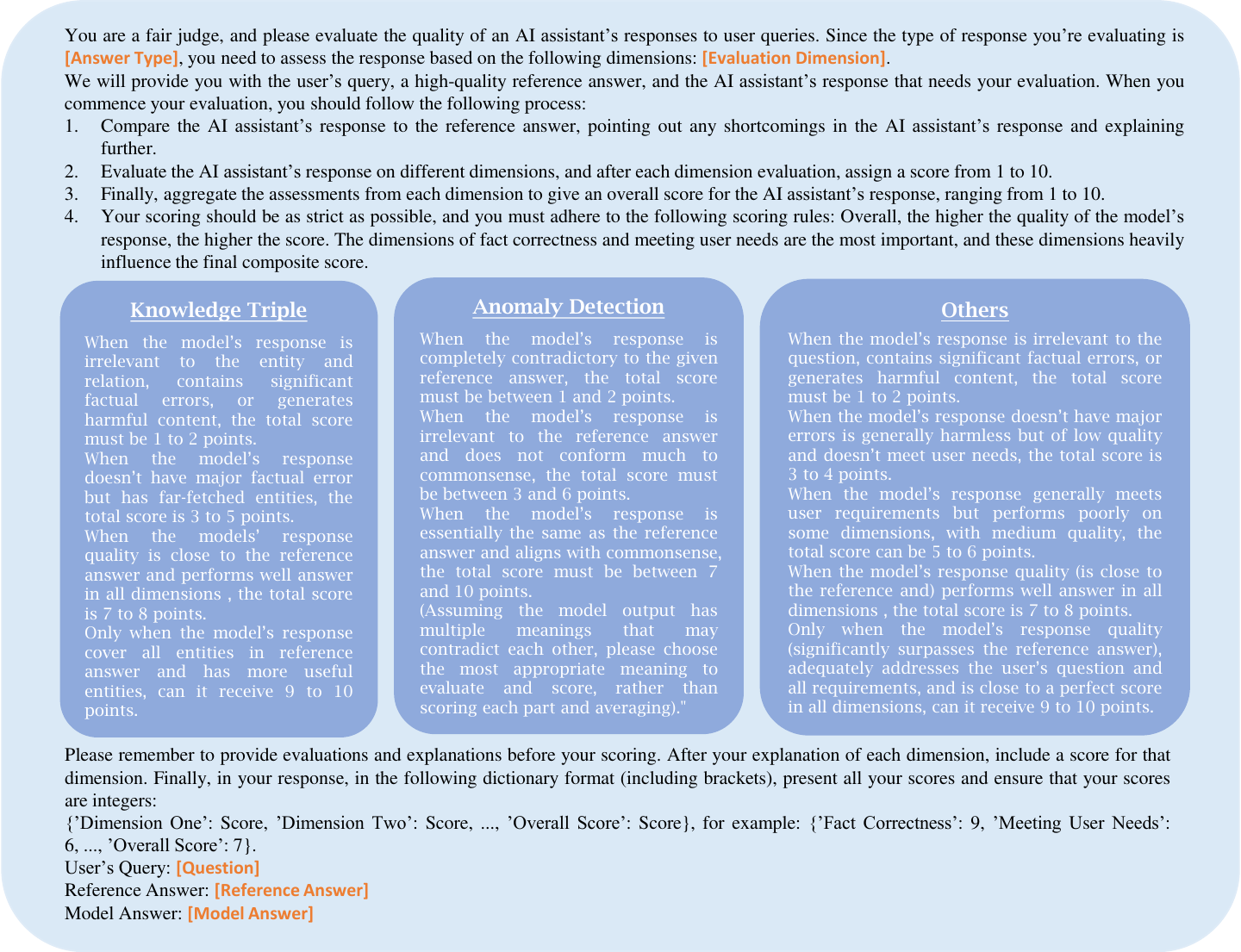}
\caption{
    The prompt template of API Evaluation following \citet{liu2023alignbench}. The \textcolor{my_orange}{\textbf{orange}} sections enclosed in brackets represent the evaluation dimensions and evaluation subjects defined according to different sub-datasets. The middle \textcolor{my_blue}{dark blue} section displays the different scoring criteria used by different sub-datasets. 
}
\label{fig:api_eval}
\end{figure*}

\subsection{API Evaluations}

It is widely recognized that the design of prompts is crucial for the quality of the LLM's output. We follow AlignBench~\cite{liu2023alignbench} to use the multi-dimensional rule-calibrated LLM-as-Judge as our evaluation prompt, and make some adjustments to the details to accommodate the tasks we designed. Specifically, all sub-datasets that use API Evaluation share the same scoring process, output requirements, and example input. However, specific evaluation strategies vary according to the different datasets. Detailed evaluation prompt template is shown in Figure \ref{fig:api_eval}. 

\subsection{Results Normalization}
\label{sec:norm_res_appendix}

As mentioned in Equation \ref{eq:norm}, we introduce a self-adaptive normalization method, aiming to scale the scores in each sub-dataset into reasonable ones. Since Equation \ref{eq:norm} is a form of linear scaling, we will first focus on the naive form before introducing our self-adaptive method. When using naive linear scaling, $\alpha$ and $\beta$ are calculated by the following equation:
\begin{equation}
    \begin{aligned}
        \alpha &= \mathbf{s}^{rank}_{1},\\
        \beta &= \mathbf{s}^{rank}_{-1},
    \end{aligned}
\end{equation}
where $\mathbf{s}^{rank}_{i}$ and $\mathbf{s}^{rank}_{-i}$ refer to the i-th highest and the i-th lowest scores among all LLMs.

To make our results more reasonable and meaningful, we make the following considerations to replace the direct use of maximum and minimum values:
\begin{itemize}
    \item Remove the boundary scores. In statistics, to enhance the stability of results and avoid outliers, it is common to remove the highest and lowest scores. We follow this approach and select the second highest $\mathbf{s}^{rank}_{2}$ and the second lowest scores $\mathbf{s}^{rank}_{-2}$ .
    \item Compress the range of scores. Considering the possibility that models better or worse than these 13 LLMs might emerge, we compress the scores into the range of 10 to 90, instead of the full 0 to 100.
\end{itemize}

Therefore, in our self-adaptive methods, $\alpha$ and $\beta$ are calculated by the following equation:
\begin{equation}
    \begin{aligned}
        \alpha &= \frac{\mathbf{s}^{rank}_{2}}{0.9}, \\
        \beta &= \alpha - \frac{\mathbf{s}^{rank}_{2} - \mathbf{s}^{rank}_{-2}}{0.8}.
    \end{aligned}
\end{equation}

Besides, $\gamma$ controls the range of the final score. The influence of the value of $\gamma$ is shown in Table~\ref{tab:gamma}. The results show that the difference in correlation for different values is not significant. The average correlation coefficient for $\gamma$ values of 1 and 2.5 are all the best. However, we observe that when the score is set to 1, the difference in scores between models is too small, leading to reduced distinction in the results. Therefore, F-Eval chooses a $\gamma$ value of 2.5.

\begin{table*}[htb]
\renewcommand\arraystretch{1.1}
  \centering
  \small
  \setlength{\tabcolsep}{3pt}
  \tabcolsep=0.3cm
  \begin{tabular}{@{}lcccccccc@{}}
    \toprule
        \multirow{2}{*}{Metrics}  & \multicolumn{2}{c}{Expression} & \multicolumn{2}{c}{Commonsense} & \multicolumn{2}{c}{Logic} & \multicolumn{2}{c}{Average}      \\
        \cmidrule(l{8pt}r{8pt}){2-3} \cmidrule(l{8pt}r{8pt}){4-5} \cmidrule(l{8pt}r{8pt}){6-7} \cmidrule(l{8pt}r{8pt}){8-9}
        & $r$ & $\rho$ & $r$ & $\rho$ & $r$ & $\rho$ & $r$ & $\rho$  \\
        \midrule
         Rank standard  & 0.242 & 0.286 & 0.706 & 0.673 & 0.432 & 0.38 & 0.46 & 0.446\\
        Self-adaptive ($\gamma=1$) & \textbf{0.767} & \textbf{0.764} & \textbf{0.924}  & \textbf{0.904} & 0.696 & \textbf{0.557} & 0.796 & \textbf{0.742} \\
        Self-adaptive ($\gamma=2.5$) & \textbf{0.767} & \textbf{0.764} & 0.918  & \textbf{0.904} & 0.706 & \textbf{0.557} & \textbf{0.797} & \textbf{0.742} \\
        Self-adaptive ($\gamma=3.5$) & 0.766 & \textbf{0.764} & 0.91  & \textbf{0.904} & \textbf{0.712} & \textbf{0.557} & 0.796 & \textbf{0.742} \\
        Self-adaptive ($\gamma=5$) & 0.765 & \textbf{0.764} & 0.897  & \textbf{0.904} & 0.707 & \textbf{0.557} & 0.79 & \textbf{0.742} \\
         
    \bottomrule
  \end{tabular}
  \caption{Comparison of Pearson ($r$) and Spearman ($\rho$) correlation coefficients in expression, commonsense and logic dimensions with different hyper-parameter $\gamma$ mentioned in Equation \ref{eq:norm}.}
  \label{tab:gamma}
\end{table*}

\section{Models being Evaluated}
\label{sec:appendix_models}

\paragraph{Llama2} Llama2~\cite{touvron2023llama} is a collection of pre-trained and fine-tuned generative text models ranging in scale from 7 billion to 70 billion parameters. We choose two base models: Llama2-7B and Llama2-13B.
\paragraph{Baichuan2} Baichuan2~\cite{baichuan2023baichuan2} is the new generation of large-scale open-source language models launched by Baichuan Intelligence Incorporated. It is trained on a high-quality corpus with 2.6 trillion tokens and has achieved the best performance in authoritative Chinese and English benchmarks of the same size. We choose two base models: Baichuan2-7B and Baichuan2-13B.
\paragraph{Qwen} Qwen~\cite{qwen} is proposed by Alibaba Cloud. It is a Transformer-based large language model, which is pre-trained on a large volume of data, such as web texts, books, codes. We choose three base models: Qwen-7B, Qwen-14B and Qwen-72B.
\paragraph{ChatGLM} ChatGLM series models are a series of dialogue pre-training models jointly released by ZhiPu AI and Tsinghua University's KEG Lab, with the aim of improving the fluency, intelligence, and diversity of dialogue. The versions we test are ChatGLM2~\cite{du2022glm} and ChatGLM3-base~\cite{zeng2022glm}.
\paragraph{DeepSeek} DeepSeek~\cite{DeepSeek-llm} is a large language model independently developed by DeepSeek, an artificial intelligence company under High-Flyer Quantitative. DeepSeek has been trained from scratch on a vast dataset of 2 trillion tokens in both English and Chinese. We choose three base models: DeepSeek-7B and DeepSeek-67B.
\paragraph{GPT} GPT series are LLMs from OpenAI, which is improved through human feedback-driven reinforcement learning to be more compliant with human instructions, more useful, harmless, and honest. Among them, GPT4.0~\cite{openai2023gpt4} is currently the most powerful model on the market, supporting image input, and it has gone through a well-designed post-training alignment process, making it larger in scale than most existing models. GPT4.0 has achieved human-level performance in various benchmark tests and even achieved top 10\% scores in some simulated exams. Here, we tested two versions: GPT3.5~\cite{chatgpt} and GPT4.0.

\section{Complete Results}
\label{sec:results_appendix}

The normalized complete results of each sub-dataset are shown in Table \ref{tab:expression}, Table \ref{tab:commonsense} and Table~\ref{tab:logic}.

\begin{table*}[!ht]
\renewcommand\arraystretch{1.2}
  \setlength{\tabcolsep}{5pt}
  \centering
  \small
  \begin{tabular}{lccccc}
  \toprule
    \textbf{Model} & \textbf{Word Diversity} & \textbf{Informative} & \textbf{Rule Following} & \textbf{Emotion Consistency} & \textbf{Average} \\
        \midrule
        Llama2-7B  &9.64 &22.43  &21.56 & 40.93 &23.64 \\
        Llama2-13B  & 20.31& 36.4 &23.75 & 54.61 &33.77 \\
        Baichuan2-7B	&47.19 & 63.51 & 48.96&31.42&47.77   \\
        Baichuan2-13B	&43.97 & 67.26 & 48.96&42.61& 50.7  \\
        Qwen-7B &49.5 & 53.01 & 53.84 &27.14 & 43.31 \\
        Qwen-14B	&47.63 & 69.58 &55.19 &36.04&52.11   \\
        Qwen-72B	& 47.66& 69.37 & 48.96&28.52& 48.63  \\
        ChatGLM2	&42.16 & 44.85 & 23.75&18.69 &32.36  \\
        ChatGLM3-base	&46.61 & 34.83 & 42.76& 23.25 &36.86 \\
        DeepSeek-7B	&35.04 & 50.19 &48.96 &64.53 & 49.68  \\
        DeepSeek-67B	&51.35& 66.61 & 64.18 &69.12 & 62.82  \\
        GPT3.5	& 65.31& 83.44 &69.71 & 58 &69.12 \\
        GPT4.0 &70.93 & 79.8 &92.12 &70.57 & 78.36 \\
  \bottomrule
  \end{tabular}
  \caption{Experiment results of expression dimensions.} \label{tab:expression}
\end{table*}

\begin{table*}[!ht]
\renewcommand\arraystretch{1.2}
  \setlength{\tabcolsep}{5pt}
  \centering
  \small
  \begin{tabular}{lcccccc}
  \toprule
    \textbf{Model} & \textbf{Commonsense Triple} & \textbf{CommonsenseQA} & \textbf{TextbookQA} & \textbf{Story} & \textbf{Instruction}& \textbf{Average} \\
        \midrule
        Llama2-7B  &26.59 &  21.98& 21.93 & 22.65 & 24.55 & 23.54 \\
        Llama2-13B  &33.95 & 24.3 & 20.06 &26.53  &26.25 & 26.22 \\
        Baichuan2-7B	&24.91 & 27.63 & 20.98 &31.8&  25.72 & 26.21 \\
        Baichuan2-13B	&24.54 & 32.19 & 36.56 &29.82& 25.87 & 29.8 \\
        Qwen-7B & 34.71 &  38.13& 28.55 &30.8 & 47.91 & 36.02 \\
        Qwen-14B	& 45.33 & 46.59 & 38.21 &32.82&  57.61 & 44.11  \\
        Qwen-72B	& 42.35 & 48.76 &49.18 & 54.52 & 59.32 & 47.66 \\
        ChatGLM2	& 38.78 & 25.93 & 18.53 &24.75 & 42.69 & 30.14 \\
        ChatGLM3-base	& 27.38 & 34.11 & 22.91 &28.85  & 51.03 & 32.86 \\
        DeepSeek-7B	& 23.68 &36.09 &23 & 32.12 &  27.25 & 29.99 \\
        DeepSeek-67B	& 40.06 & 41.25& 53.47 &32.82 & 29.22  & 39.36 \\
        GPT3.5	& 70.61 & 70.34 & 64.97 &71.33  & 71.9 & 69.83 \\
        GPT4.0 & 76.85 & 77.81 & 83.45 &70.85 & 73.38 & 76.47 \\
  \bottomrule
  \end{tabular}
  \caption{Experiment results of commonsense dimensions.} \label{tab:commonsense}
\end{table*}

\begin{table*}[!ht]
\renewcommand\arraystretch{1.2}
  \setlength{\tabcolsep}{5pt}
  \centering
  \small
  \begin{tabular}{lcccccccc}
  \toprule
    \textbf{Model} & \textbf{~~~~ICL~~~~} & \textbf{~~~~COT~~~~} & \textbf{Fallacy Attack} & \textbf{Contradiction} & \textbf{Coreference} & \textbf{Anomaly Detection} & \textbf{Average} \\
        \midrule
        Llama2-7B  & 21.75&  9.36& 31.32  & 37.39& 22.71& 20.67& 26.63  \\
        Llama2-13B  & 35.94& 30.36 & 28.51 & 53.77& 22.74& 14.51&36.12   \\
        Baichuan2-7B	& 24.05& 23.98 & 31.82 &45.46 &21.29 &  14.51&  28.71   \\
        Baichuan2-13B	&63.88 & 43.02 & 29.82 & 64.47& 32.65& 9.96& 37.69   \\
        Qwen-7B &38.38 & 35.66 & 33.74  & 40.03& 18.66& 16.35&  28.66   \\
        Qwen-14B	&58.26 & 54.24&40.94  &51 &58.93 & 18.46& 47.96    \\
        Qwen-72B	& 67.26& 64.79 &37.27  &45.46 & 48.58& 44.98&52.07     \\
        ChatGLM2	& 7.53&  32.08&22.19  &66.97 & 9.06& 20.67& 23.33    \\
        ChatGLM3-base	& 46.95&  61.06& 23.49 & 59.23& 27.54&12.87 &38.06    \\
        DeepSeek-7B	&40.06 &  19.94& 44.87 & 42.72& 19.92& 8.84& 31.82    \\
        DeepSeek-67B	& 66.47& 62.01 & 46.45 & 37.39& 62.89& 44.98& 54.95   \\
        GPT3.5	&75.78 &  64.79& 69.41  & 86.03& 36.51&38.16 &  62.25  \\
        GPT4.0& 59.97& 76.31 & 80.94 & 81.53& 75.25& 98.57&80.22   \\

  \bottomrule
  \end{tabular}
  \caption{Experiment results of logic dimensions.} \label{tab:logic}
\end{table*}

\section{Human Annotation}
\label{sec:anno}

Three primary tasks require manual execution in our paper: constructing some challenged dataset, scoring the model results, and providing reference answers for reference-free subjective tasks. We assemble a total of 10 annotators, comprising 3 experts in the fields of linguistics and NLP, and 6 additional annotators who have passed our preliminary testing. 6 ordinary annotators are grouped in pairs, each responsible for one of three tasks. Within each group, every annotator is instructed to independently complete data construction or annotation tasks for further cross validation. Meanwhile, the 3 experts are involved in the review process for all three tasks, randomly examining the quality of 50\% data. The final overall failure rate averages 2.15\%. Specifically, for annotation tasks where the results of the two annotators significantly diverge, experts will conduct a focused review and unify the final outcome. The detailed annotation process for each task is as follows.

\subsection{Data Collection on Fallacy Attack}
The crux of the Fallacy Attack dataset involves the creation of a syllogism. It includes two premises that maintain a connection yet are distinctly different, with the ultimate aim of prompting the model to produce counter-intuitive statements. Initially, two annotators create a variety of syllogisms across different domains, ensuring that the first premise aligns with commonsense and the second premise retains a link to the first. Based on these initial examples, GPT4.0 generates a wider array of examples, from which those that meet our requirements are selected to form the preliminary dataset.

An expert then reviews this data, providing commonsense answers for each syllogism. This stage is crucial for identifying and reshaping examples that are ambiguous or lack clear common-sense reasoning. These expert-provided common-sense responses are integrated into the dataset and serve as part of the prompts for model inference.

\begin{table*}[!t]
\renewcommand\arraystretch{1.1}
  \centering
  \small
  \setlength{\tabcolsep}{3pt}
  \tabcolsep=0.3cm
  \begin{tabular}{@{}lcccccccc@{}}
    \toprule
        \multirow{2}{*}{Metrics}  & \multicolumn{2}{c}{Expression} & \multicolumn{2}{c}{Commonsense} & \multicolumn{2}{c}{Logic} & \multicolumn{2}{c}{Average}      \\
        \cmidrule(l{8pt}r{8pt}){2-3} \cmidrule(l{8pt}r{8pt}){4-5} \cmidrule(l{8pt}r{8pt}){6-7} \cmidrule(l{8pt}r{8pt}){8-9}
        & $r$ & $\rho$ & $r$ & $\rho$ & $r$ & $\rho$ & $r$ & $\rho$  \\
        \midrule
         Annotator 1 \& 2  & 0.848 & 0.779 & 0.968 & 0.989 & 0.854 & 0.85 & 0.890 & 0.873 \\
        Annotator 2 \& 3 & 0.956 & 0.968 &  0.786 & 0.879 & 0.708 &0.834 & 0.817 & 0.894 \\
        Annotator 1 \& 3 & 0.684 & 0.692 & 0.723 & 0.873 & 0.634 & 0.676 & 0.680 & 0.747 \\
        Average & 0.829 & 0.813 &  0.826 & 0.914 & 0.732 & 0.787 & 0.796 & 0.838 \\
         
    \bottomrule
  \end{tabular}
  \caption{Human agreement of the annotation for meta evaluation in expression, commonsense and logic dimensions.}
  \label{tab:human}
\end{table*}

\subsection{Human Scoring for Meta Evaluation}

In order to compute the correlation of our evaluation methods with human judgements, we need to gather human evaluation scores for responses provided by different LLMs. To enhance annotation efficiency, we upload generations from all LLMs on each sub-dataset to the LabelU annotation platform\footnote{https://labelu.shlab.tech/} in batches. The 3 annotators are then assigned to simultaneously rate these generations on a scale of 0 to 10. Upon receiving the scores, we automatically identify generations where the ratings of the three annotators vary significantly ($\geq$5) and pass these on to the responsible expert to determine the final score. To ensure the consistency of the scores assigned by our chosen annotators, we additionally enlist one more annotator to label a sub-dataset in each of the three dimensions. We calculate the Pearson ($r$) and Spearman ($\rho$) correlation scores of the three sub-datasets across the three annotators. As shown in Table \ref{tab:human}, it can be seen that the correlation scores in each dimension are relatively high, demonstrating that the scores from the annotators are quite reliable.

\subsection{Annotation for Reference-free Sub-datasets}

When computing BLEU and BERTScore, the dataset requires reference answers to calculate the similarity between outputs and answers. Therefore, we need to manually annotate reference answers for reference-free subjective sub-datasets. We add detailed instruction requirements to the questions in the Word Diversity, Informative, Emotion Consistency, and Contradiction sub-datasets. Then we post all questions on the LabelU platform, dividing them into two parts for two annotators. In the annotation progress, we randomly select 60\% of the responses for expert review. Responses of low quality are reassigned to another annotator for re-answering. This process is repeated until all sampled responses meet the required standards.

\end{document}